\documentclass[10pt,journal,compsoc]{IEEEtran}

\ifCLASSOPTIONcompsoc
  \usepackage[nocompress]{cite}
\else
  \usepackage{cite}
\fi

\ifCLASSINFOpdf
  \usepackage[pdftex]{graphicx}
  
\else
  
\fi

\usepackage{amsmath}

\usepackage{algorithmic}

\ifCLASSOPTIONcompsoc
 \usepackage[caption=false,font=footnotesize,labelfont=sf,textfont=sf]{subfig}
\else
 \usepackage[caption=false,font=footnotesize]{subfig}
\fi

\usepackage{algorithm}
\usepackage{multirow}
\usepackage{makecell}
\usepackage{booktabs}
\usepackage{marvosym}
\usepackage{pifont}
\usepackage{xcolor}
\usepackage{hyperref}
\usepackage{amssymb}
\usepackage{times}

\newcommand{\ymark}{\text{\ding{51}}}

\newcommand{\myparagraph}[1]{\vspace{5pt} \noindent \textbf{#1.}}

\begin{document}

\title{Omni-Training: Bridging Pre-Training and Meta-Training for Few-Shot Learning}

\author{Yang~Shu,
        Zhangjie~Cao,
        Jinghan~Gao, 
        Jianmin~Wang,
        Philip~S.~Yu,~\IEEEmembership{Fellow,~IEEE},
        Mingsheng~Long,~\IEEEmembership{Member,~IEEE}%
\IEEEcompsocitemizethanks{
\IEEEcompsocthanksitem The authors are with the School of Software, BNRist, Tsinghua University, Beijing 100084, China.
\IEEEcompsocthanksitem Corresponding author: Mingsheng Long, mingsheng@tsinghua.edu.cn.
}}

%\markboth{IEEE Transactions on Pattern Analysis and Machine Intelligence,~Vol.~XX, No.~X}%
%{Shu \MakeLowercase{\textit{et al.}}: Omni-Training: Bridging Pre-Training and Meta-Training for Few-Shot Learning}

\IEEEtitleabstractindextext{%
\begin{abstract}
Few-shot learning aims to fast adapt a deep model from a few examples. While pre-training and meta-training can create deep models powerful for few-shot generalization, we find that pre-training and meta-training focuses respectively on cross-domain transferability and cross-task transferability, which restricts their data efficiency in the entangled settings of domain shift and task shift. We thus propose the Omni-Training framework to seamlessly bridge pre-training and meta-training for data-efficient few-shot learning. Our first contribution is a tri-flow Omni-Net architecture. Besides the joint representation flow, Omni-Net introduces two parallel flows for pre-training and meta-training, responsible for improving domain transferability and task transferability respectively. Omni-Net further coordinates the parallel flows by routing their representations via the joint-flow, enabling knowledge transfer across flows. Our second contribution is the Omni-Loss, which introduces a self-distillation strategy separately on the pre-training and meta-training objectives for boosting knowledge transfer throughout different training stages. Omni-Training is a general framework to accommodate many existing algorithms. Evaluations justify that our single framework consistently and clearly outperforms the individual state-of-the-art methods on both cross-task and cross-domain settings in a variety of classification, regression and reinforcement learning problems.
\end{abstract}

}

\maketitle

\IEEEdisplaynontitleabstractindextext

\IEEEpeerreviewmaketitle

\IEEEraisesectionheading{\section{Introduction}\label{sec:introduction}}

\IEEEPARstart{D}{eep} learning~\cite{cite:Nature2015DeepLearning} has achieved the state-of-the-art performance in various machine learning tasks~\cite{cite:CVPR2016ResNet,silver2016mastering,devlin2018bert,adiwardana2020towards}. However, most deep learning methods, in particular the foundation models \cite{bommasani2021opportunities}, are ``data hungry'', in that the success of these methods highly relies on large amounts of labeled data. This clearly limits the application of deep learning to widespread domains or tasks, especially those with sparse data and insufficient annotations, such as personalized healthcare~\cite{yu2019reinforcement}. In order to promote the grounding of deep learning models, few-shot learning, which aims to fast learn various complex tasks from a few labeled data, has attracted enormous attention recently~\cite{cite:TPAMI2006FewShot, cite:ICML2017MAML,wang2021self}.

\begin{figure*}[t]
	\centering
	\includegraphics[width=0.96\textwidth]{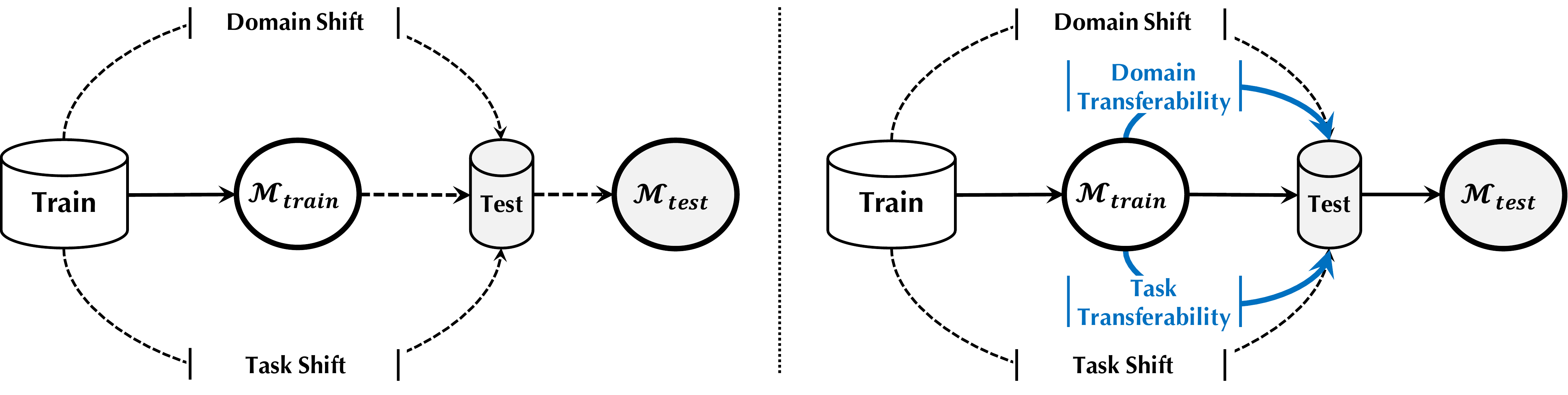}
	\caption{Illustration of the two challenges of few-shot learning. Due to the \emph{domain shift} and \emph{task shift} between the training dataset and the test dataset, it is hard for the trained model $\mathcal{M}_\texttt{train}$ to transfer to the test set and boost its data-efficiency. An ideal training method should learn representations with both \emph{domain transferability} and \emph{task transferability} and adapt $\mathcal{M}_\texttt{train}$ to downstream model $\mathcal{M}_\texttt{test}$ in a data-efficient way.}
	\label{fig:figure1}
\end{figure*}

Human beings are gifted with the ability to quickly learn new tasks by making use of previous experience and knowledge. In analogy to this, deep learning models can reuse the representations learned previously to help efficiently solve widespread downstream tasks. Recent advances have revealed that a properly trained model endows an important property: \emph{transferability}, and higher transferability indicates better generalizability to new scenarios. In general situations as illustrated by Figure~\ref{fig:figure1}, complex relationships between the pretext dataset and the new task hinder the downstream learning and pose challenges to the transferability of learned representations. The two main challenges come from the different distributions across domains, \emph{i.e.}~domain shift and different semantics across tasks, \emph{i.e.}~task shift. For example, in image classification, different domains may have different visual factors such as different styles, viewpoints and lighting, while different tasks may have different categories. In most cases, the two challenges entangle with each other, making few-shot learning a very hard problem. Thus, a versatile algorithm should bridge these two gaps and learn representations with both \emph{domain transferability} and \emph{task transferability}.

Two mainstream representation learning paradigms for few-shot learning are \emph{pre-training} and \emph{meta-training}. In pre-training, we train a high-capacity model for a pretext task on large-scale datasets ~\cite{devlin2018bert,kolesnikov2020big} and fine-tune the model on the target task~\cite{cite:GPT}. In meta-training, we train the model from diverse tasks and fast adapt the model to new tasks~\cite{cite:NIPS2016MatchingNet,cite:ICML2017MAML,cite:NIPS2017ProtoNet}. As evidenced by recent studies, neither paradigm can dominate in the widespread few-shot learning scenarios~\cite{cite:ICLR2019ACloserLook,cite:ECCV2020BroaderCrossDomain,cite:Arxiv2021ComparingTransferMeta}, because it requires knowledge that generalizes across both domains and tasks. Pre-training representations can transfer to widespread domains, since the pretext task is designed to be general across domains. However, only pre-training on a single pretext task makes it hard to fast adapt to many new tasks. In contrast, the diverse tasks equip meta-training with the ability to fast adapt across many tasks with extremely sparse data, but the meta-training tasks are usually domain-specific and thus the learned representations cannot generalize well across domains.

In line with the understanding of pre-training and meta-training, we further study both paradigms with regard to the two transferability properties and reach a similar conclusion: pre-training methods are apt at the domain transferability while meta-training methods at the task transferability. We then take a step forward to exploit the collaboration between pre-training and meta-training and draw an important finding that neither a simple ensemble nor a tight combination can achieve both kinds of transferability. This finding motivates us to design a new Omni-Training framework to bridge both sides for few-shot learning.

Omni-Training seamlessly bridges pre-training and meta-training to learn deep representations with both domain transferability and task transferability. The first part is Omni-Net, a tri-flow architecture. Besides a joint-flow for shared representation learning, Omni-Net introduces two new parallel flows for pre-training and meta-training to yield representations of domain transferability and task transferability respectively. It further coordinates the parallel flows by routing their representations via the joint-flow, making each gain the other kind of transferability. The second part is Omni-Loss, which works in cooperation with the architecture for learning transferable representations. A self-distillation strategy is imposed to both the pre-training and meta-training objectives, forcing the parallel flows to learn more transferable representations. Omni-Training is a general framework that can accommodate many existing pre-training and meta-training algorithms. Thorough evaluations on cross-task and cross-domain datasets in classification, regression and reinforcement learning problems show that Omni-Training consistently and clearly outperforms the individual state-of-the-art deep learning methods.

\section{Related Work}

Few-shot learning aims to make full use of every sample and address new tasks with a few labeled data~\cite{cite:TPAMI2006FewShot,wang2020generalizing,yu2018towards}. In this paper, we focus on representation learning algorithms towards it, which aim to learn transferable representations from pretext data to reduce the data requirement of learning new tasks. We restrict our review to two mainstream categories of representation learning algorithms for few-shot learning that achieve state-of-the-art performance: pre-training and meta-training.

\subsection{Pre-Training}

One line of few-shot learning methods is to learn deep representations by pre-training deep networks with a pretext task on the training datasets. With the prevalence of large-scale labeled datasets and the advanced computational infrastructure, deep networks with extremely big model capacity are trained for various applications such as computer vision~\cite{szegedy2016rethinking,he2019rethinking,kolesnikov2020big} and natural language processing~\cite{devlin2018bert,radford2019language}. With such deep models, recent works re-take the pre-training and fine-tuning paradigm and demonstrate that fine-tuning high-capacity deep models pre-trained on large datasets achieves state-of-the-art performance in various applications with only a few labeled data~\cite{cite:NIPS2020LanguageModelFewShot,cao2021transfer,zhu2020transfer}. Pre-training is also adopted in reinforcement learning to enable learning the policy for new environments with less interaction steps~\cite{xie2021policy,campos2021beyond,schwarzer2021pretraining}. More advanced pre-training strategies also boost few-shot learning performance, such as training an ensemble of models~\cite{cite:ICCV2019EnsembleFewShot} and training with knowledge distillation~\cite{cite:ECCV2020RethinkingFewShot}.

There are methods towards the stage of fine-tuning on the new task. For example, some works reuse the representations to predict parameters of new categories~\cite{cite:CVPR2018PredictingParameters,qi2018low}. Some works regularize the model of the new task from the aspects of parameters or representations to fully extract the knowledge of the pre-trained models~\cite{cite:ICML18L2SP, cite:ICLR19Delta}. Recent research also proposed to explore relationships between the training and test datasets and mitigate negative transfer~\cite{cite:NIPS19BSS, cite:NIPS20CoTuning}. Cao~\textit{et al.}~\cite{cao2021transfer} proposed an ease-in-ease-out fine-tuning method to enable transfer reinforcement learning across homotopy classes. These methods focus on a different perspective and are in parallel with this paper.

Pre-training approaches are simple and effective to improve data efficiency in new scenarios, which show higher domain transferability and outperform sophisticated meta-training methods in the cross-domain setting~\cite{cite:ICLR2019ACloserLook,cite:ECCV2020BroaderCrossDomain,cite:Arxiv2021ComparingTransferMeta}. However, as the training stage only involves one pretext task, these methods cannot quickly handle the rapid changes of semantics in new tasks~\cite{cite:ICML2017MAML}.

\subsection{Meta-Training}

Meta-training addresses few-shot learning by learning representations generalizable across many training tasks, which can be naturally adapted to new tasks~\cite{cite:1987EvolutionaryPrinciples, cite:IJCNN1992MetaNetworks, cite:LearningtoLearnBook}. It has been widely used in a variety of applications.

Few-shot learning \cite{cite:TPAMI2006FewShot} is widely studied in the field of classification, especially image recognition, where a typical form is to learn from a few annotated data, \emph{i.e.} the ``N-way-K-shot'' few-shot classification problems~\cite{wang2020generalizing,cite:Arxiv20FewShotSurvey}. Metric-based meta-learning methods are tailored for these problems, which learn an embedding space to form decision boundaries according to the distances between samples~\cite{koch2015siamese,cite:NIPS2016MatchingNet,cite:NIPS2017ProtoNet,cite:CVPR2018RelationNet,cite:ICML2019IMP}. Recently, embedding functions are improved by stronger inductive bias such as graph networks~\cite{cite:ICLR2018GNN}, fine-grained attention maps~\cite{cite:NIPS2019CrossAttention}, task-adaptive projections~\cite{cite:NIPS2018TADAM,cite:ICML2019TapNet} and set-to-set functions~\cite{cite:CVPR2020Set2Set}. 

Some other meta-learning methods deal with various applications. Early works build meta-learners to learn how to update the model parameters and generalize the updating rules to new tasks~\cite{cite:1990SynapticLearning,cite:NeuralComputation1992LearningtoControl}, which have been recently applied in deep learning to enable fast adaptation of deep networks~\cite{cite:NIPS2016LearningtoLearn,cite:ICLR2017OptimizationasaModel,cite:Arxiv2017MetaSGD}. Such learning to learn paradigm is also demonstrated to work for regression~\cite{cite:NIPS2016LearningtoLearn,cite:Arxiv2017MetaSGD} and reinforcement learning~\cite{xu2018meta,houthooft2018evolved}. Several works equip networks with external or internal memory so that meta-knowledge can be effectively stored and queried for data-efficient adaptation to new tasks~\cite{cite:ICML2016MetaLearningMemory,cite:ICML2017MetaNetworks,cite:ICLR2018NeuralAttentiveMeta,cite:ICML2018ConditionallyShiftedNeuron}. The memory-augmented models are also applied to reinforcement learning to improve data-efficiency~\cite{cite:Arxiv2016FastRL,cite:Arxiv2016LearningtoRL,cite:ICLR2018NeuralAttentiveMeta}. These methods introduce additional parameters and storage costs or require a particular architecture of the learner for meta-learning.

Model agnostic meta-learning introduces the gradient-based idea, which trains a good initialization of the deep network as the meta-knowledge such that a small number of gradient steps and interactions in the new environment can induce high generalization performance~\cite{cite:ICML2017MAML}. The idea is later improved by new architectures ~\cite{cite:ICML2018GradientBased,cite:ICML2019Hierarchically}. Such gradient-based meta-training methods show strong performance in real robotics applications such as imitation learning~\cite{cite:NIPS17OneShotImitationLearning, cite:CRL17OneShotVisualImitation}, locomotion~\cite{cite:Arxiv17MetaSharedHierarchies}, visual navigation~\cite{cite:CVPR19TaskAgnosticMeta}, and robot manipulation~\cite{cite:CRL18FewShotGoalInference}. They can also be extended to other applications such as regression and image classification by changing the architecture and training objective~\cite{cite:ICML2017MAML,cite:ICLR2019LEO,cite:ICLR2019R2D2,cite:CVPR2019MetaOpt}.

Though meta-training empowers the deep representations with the ability to generalize across new tasks, a recent empirical study has revealed that meta-trained representations cannot generalize across domains with distribution shift~\cite{cite:ICLR2019ACloserLook}. Tseng~\textit{et al.}~\cite{cite:ICLR2020CrossDomain} use feature-wise transformation layers to simulate various image feature distributions extracted from the training tasks in different domains. However, the domain transferability is still limited especially in domains with large distribution shift~\cite{cite:ECCV2020BroaderCrossDomain}. Our method acquires the missing piece of domain transferability from \emph{pre-training}, which does not require multiple pretext domains but achieves better cross-domain generalization ability.

Meta-training and pre-training are apt at task transferability and domain transferability respectively, and neither can dominate the other. A natural idea is to integrate two types of approaches to achieve both. Sun~\textit{et al.}~\cite{cite:CVPR2019MetaTransferLearning} simply chain the process of pre-training and meta-training, but such a simple combination still lacks both kinds of transferability. 
In contrast, our Omni-Training framework seeks to flexibly bridge pre-training and meta-training to empower both kinds of transferability.

\section{Background and Analysis}

We first introduce few-shot learning and its two key prerequisites: \emph{domain transferability} and \emph{task transferability}. Then we delve into two mainstream methods, \emph{pre-training} and \emph{meta-training}, each of which learns a representation of a specific kind of transferability and enables generalization to either new domains or new tasks.

\subsection{Few-Shot Learning}

At the training phase, the goal is to learn a feature representation $F$ on the training set $\mathcal{D}_\texttt{train}$ of sufficient labeled examples, which enables fast solving new tasks from a few examples. At the testing phase, the learned representation $F$ is evaluated on new tasks, either within domain or across domains. Each task comes with a test set $\mathcal{D}_\texttt{test}=\{\mathcal{S}_\texttt{test}, \mathcal{Q}_\texttt{test}\} $ partitioned into a support set $\mathcal{S}_\texttt{test}$ with a few labeled examples and a query set $\mathcal{Q}_\texttt{test}$ with many unlabeled examples to predict. The learned representation $F$ should adapt fast to each new task through the support set and then yield accurate predictions on the query set. 

The key to enable few-shot learning in downstream tasks is the transferability of the representations.
Given input $\mathbf{x} \in \mathcal{X}$ and output $\mathbf{y} \in \mathcal{Y}$, denote the joint distribution as $\mathcal{P}(\mathbf{x}, \mathbf{y})$ and the learning task as $\mathcal{T}: \mathbf{x} \mapsto \mathbf{y}$.  
The \emph{domain transferability} measures the generalizability under train-test distribution shift, $\mathcal{P}_\texttt{train} \neq \mathcal{P}_\texttt{test}$, and the \emph{task transferability} measures the generalizability under train-test task shift, $\mathcal{T}_\texttt{train} \neq \mathcal{T}_\texttt{test}$. In general situations of few-shot learning, complex relationships between the training dataset and the new tasks entangle distribution shift and task shift. So we should learn representations with both domain transferability and task transferability to enable data-efficient few-shot learning.

\begin{figure*}[t]
	\centering
	\subfloat[Task Transferability]{
		\includegraphics[width=0.315\textwidth]{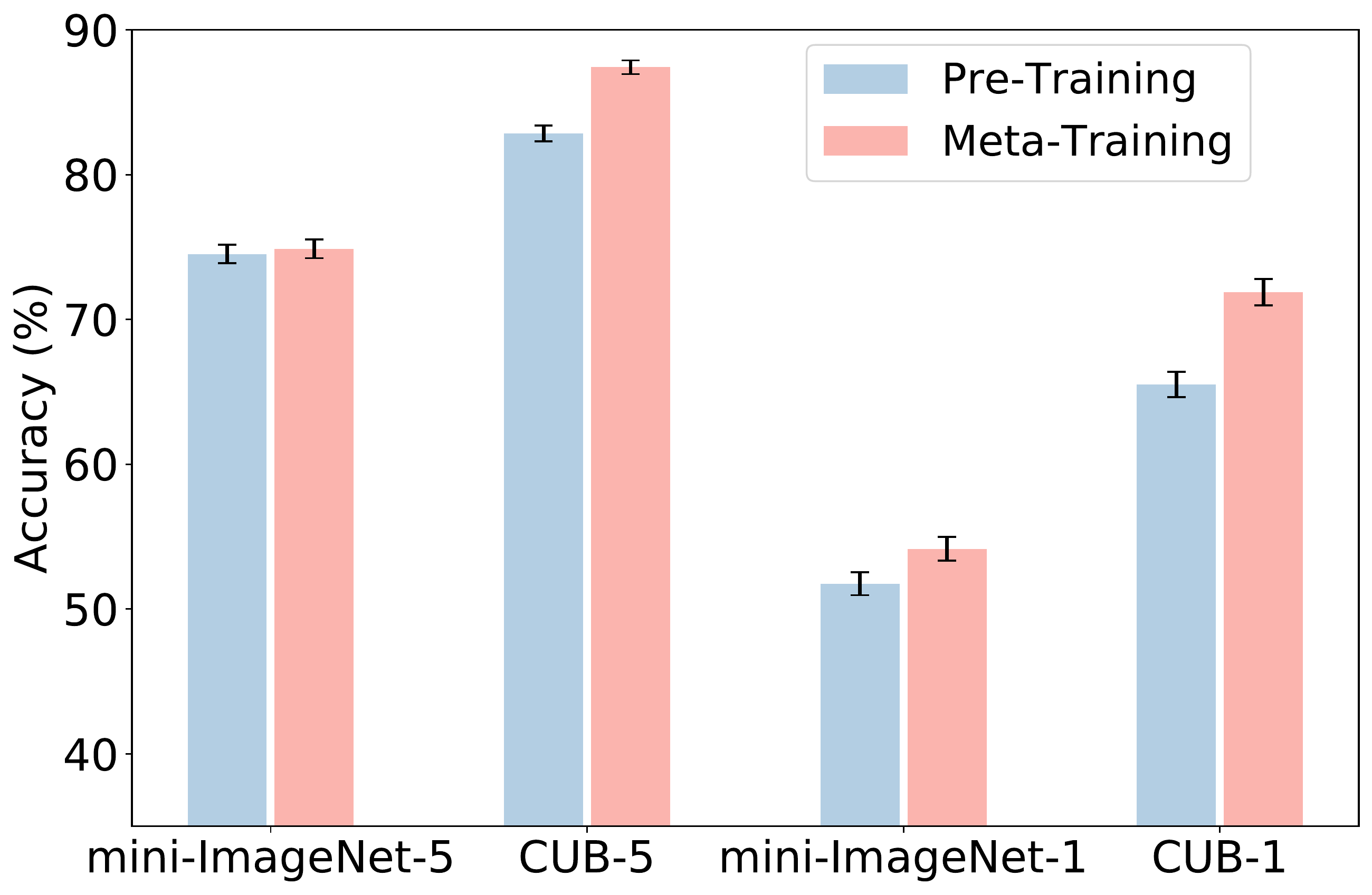}
		\label{fig:task_transferability}}
	\subfloat[Domain Transferability]{
		\includegraphics[width=0.315\textwidth]{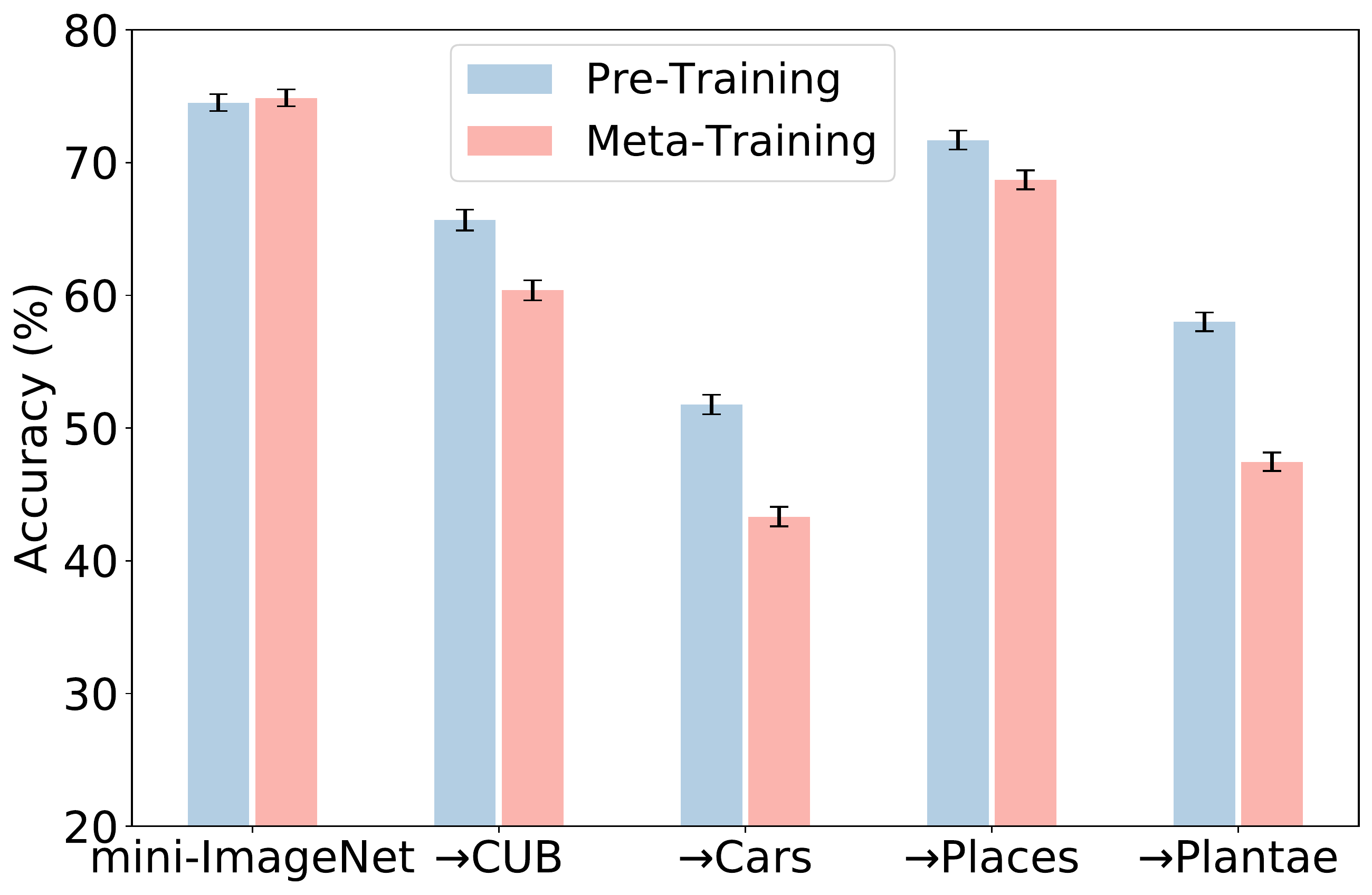}
		\label{fig:domain_transferability}}
	\subfloat[Combination Strategies]{
		\includegraphics[width=0.315\textwidth]{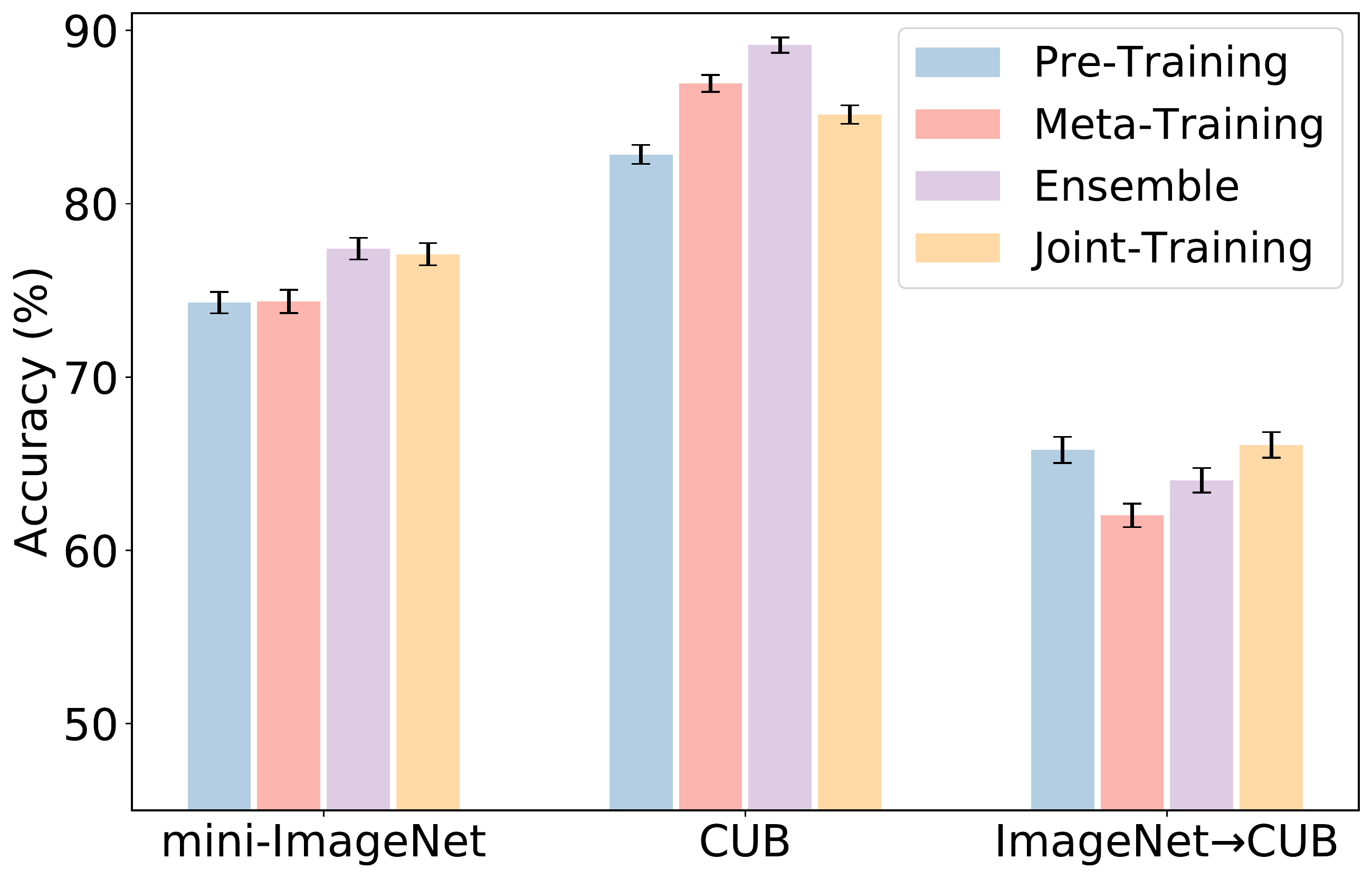}
		\label{fig:comparing_acc}}
	\hfil
	\caption{Analysis: (a) Task transferability of pre-training and meta-training in the \emph{cross-task} setting; (b) Domain transferability of pre-training and meta-training in the \emph{cross-domain} setting; (c) Accuracy of pre-training, meta-training and two combination strategies, {Ensemble} and {Joint-Training}.}
\end{figure*}

\subsection{Training Methods for Few-Shot Learning}

\myparagraph{Pre-Training} In pre-training approaches, deep representations are often learned by supervised learning on a large-scale training dataset $\mathcal{D}_\texttt{train}$, which facilitate data-efficient or few-shot learning for a variety of downstream tasks. We use an abstract model composed of a feature extractor $F$ to generate the representation and a task-specific head $H$ to predict the output, which is applicable to various tasks. 
% For example, $F$ is a convolutional network and $H$ is a classifier for a visual recognition task. 
During the training stage, the training set $\mathcal{D}_\texttt{train}$ is viewed as samples from a joint distribution of inputs and labels: $\mathcal{P}(\mathbf{x}, \mathbf{y})$. Representation learning is conducted by optimizing $H$ and $F$ over the sampled \emph{mini-batches} from the training distribution with the loss $\ell_\texttt{pre}$ tailored to the specific task or algorithm:
\begin{equation}\label{eqn:feature_learning}
	\mathop{\min}_{H,F} {\mathbb{E}}_{(\mathbf{x}, \mathbf{y}) \sim \mathcal{P}(\mathbf{x}, \mathbf{y}) } \ \ell_\texttt{pre} \big( \mathbf{y}, H \circ F \left( \mathbf{x} \right) \big).
\end{equation}
During test, we transfer the pre-trained models on the new task $\mathcal{D}_\texttt{test}=\{\mathcal{S}_\texttt{test}, \mathcal{Q}_\texttt{test}\}$. The feature extractor $F$ is fine-tuned and a task-specific head $H_\texttt{new}$ for the new task is trained with the labeled data in support set $\mathcal{S}_\texttt{test}$ and applied in query set $\mathcal{Q}_\texttt{test}$.

\myparagraph{Meta-Training} In meta-training, the representations are learned to perform well across a set of tasks sampled from a task distribution constructed from the training set. Specifically, the training set $\mathcal{D}_\texttt{train}$ is viewed as a distribution of tasks $\mathcal{T}$. Each task mimics the testing situation, which contains a support set $\mathcal{S}$ with only a few labeled samples and a query set $\mathcal{Q}$ needing predictions. The meta-learner is optimized over \emph{episodes} of tasks sampled from $\mathcal{T}$. The model $F$ and $H$ are learned to efficiently solve each of the tasks conditioned on the support set $\mathcal{S}$ with only a few samples, and updated by the performance evaluated on the query set $\mathcal{Q}$:
\begin{equation}\label{eqn:metafeature_learning}
	\mathop{\min}_{H,F} \mathbb{E}_{(\mathcal{S}, \mathcal{Q}) \sim \mathcal{P}(\mathcal{T} )} \mathbb{E}_{(\mathbf{x}, \mathbf{y}) \in \mathcal{Q}} \ \ell_\texttt{meta} \big(\mathbf{y}, H \circ F(\mathbf{x} | \mathcal{S}) \big),
\end{equation}
where $\ell_\texttt{meta}$ is the loss of specific meta-training algorithms defined on each episode, \emph{e.g.}, the meta-objective in~\cite{cite:ICML2017MAML}.
In test time, the models are fast adapted to the new task with its support set $\mathcal{S}_\texttt{test}$  in a similar way as the training phase, and the adapted models can be used for predictions on the query set $\mathcal{Q}_\texttt{test}$.

\subsection{Transferability Assessment}\label{sec:pre or meta}

We empirically compare pre-training and meta-training in terms of task transferability and domain transferability. We evaluate two typical methods, {Baseline}~\cite{cite:ICLR2019ACloserLook} as the pre-training method and {ProtoNet}~\cite{cite:NIPS2017ProtoNet} as the meta-training method. We first use two benchmarks mini-ImageNet and CUB and follow the protocol in~\cite{cite:ICLR2019ACloserLook}. Note that we use test tasks from the same dataset to relieve the influence of distribution shift and mainly focus on \emph{task shift}. As shown in Figure~\ref{fig:task_transferability}, the pre-training and meta-training methods perform comparably on mini-ImageNet-5 (5 examples per class). However, in the more extreme situation with only 1 example per class, meta-training outperforms pre-training, where the boost becomes larger on CUB: a fine-grained dataset with smaller distribution shifts between tasks. The result indicates higher task transferability of meta-training. Next, we explore the influence of \emph{distribution shift} across domains. We train the model on the mini-ImageNet dataset, but evaluate it on different domains including CUB, Cars, Places and Plantae. As shown in Figure~\ref{fig:domain_transferability}, pre-training and meta-training have similar in-domain performance, but pre-training consistently outperforms meta-training in four cross-domain situations. This result indicates higher domain transferability of pre-training.

Our key finding is that pre-training introduces higher domain transferability while meta-training introduces higher task transferability. This explains the phenomenon that both methods may fail in some few-shot learning scenarios~\cite{cite:ICML2017MAML, cite:ICLR2019ACloserLook, cite:ICLR2020MetaDataset, cite:ECCV2020BroaderCrossDomain}. In general situations, the new tasks hold complex relationships with the training set, presenting both challenges of distribution shift and task shift, which entangle with each other. For example, in the in-domain experiment, there could still be domain shift caused by different categories; In the cross-domain experiment, while domain shift is the main challenge, task transferability is still required to adapt across different classes. Overall, we need to learn representations with both domain transferability and task transferability to fully enable few-shot learning.

We study two simple ways to combine pre-training and meta-training. One is to separately train two models with two methods, and use their ensemble for prediction, denoted as {Ensemble}. The other is to jointly train the model with both training objectives, denoted as {Joint-Training}. We evaluate them on three situations of mini-ImageNet, CUB, and transferring mini-ImageNet to CUB. As shown in Figure~\ref{fig:comparing_acc}, both combination strategies promote the performance in some cases, but the improvement is minor and inconsistent. The gain of {Ensemble} indicates that pre-training and meta-training representations endow complementary knowledge. However, this simple ensemble lacks the knowledge coordination between pre-training and meta-training. The improvement of {Joint-Training} shows the importance to extract shared knowledge between the two training paradigms, but this tight combination sacrifices the specific transferability held by each approach. Such a \emph{transferability dilemma} motivates the proposed Omni-Training framework, which seeks to flexibly acquire both domain transferability and task transferability for better few-shot learning.

\section{Omni-Training Framework}\label{sec:framework}

In this paper, we are interested in learning representations with both domain transferability and task transferability by incorporating and bridging pre-training and meta-training in a unified Omni-Training framework. As discussed in Section~\ref{sec:pre or meta}, this goal is non-trivial to realize with simple combinations of these two training paradigms. Beyond the tight combination of joint-training, we have two more key insights in designing the framework. Our \textit{first key insight} is that the domain transferability of pre-training and the task transferability of meta-training should be preserved. Furthermore, there should be knowledge communication between the two types of training to enable them to complement each other. Our \textit{second key insight} is that this non-trivial unification should be realized with the design in both network architectures and training algorithms. 
These insights are embedded into the Omni-Training framework via an \textit{Omni-Net} architecture guided by an \textit{Omni-Loss}. 

\begin{figure*}[tbp]
	\centering
	\includegraphics[width=1\textwidth]{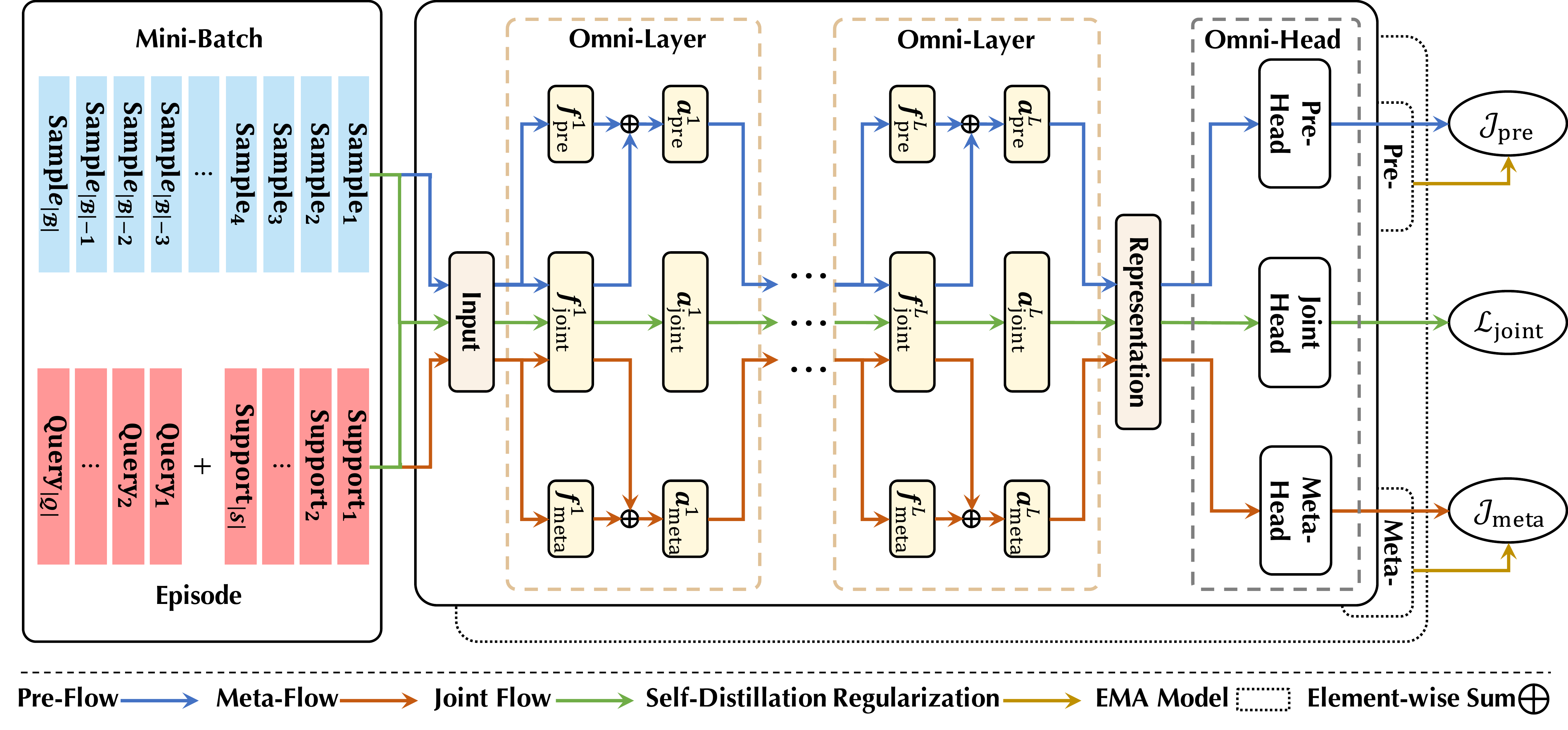}
	% \vskip 0.05in
	\vspace{-15pt}
	\caption{The Omni-Training framework consists of three data flows:  {joint-flow} (green), {pre-flow} (blue), and {meta-flow} (red). The Omni-Net consists of a backbone $F$ and an Omni-Head $H$, where $F$ is formed by stacking Omni-Layers and $H$ is formed of three heads $H_\texttt{joint}$, $H_\texttt{pre}$ and $H_\texttt{meta}$. Each Omni-Layer has a main chunk layer $f_\texttt{joint}$ and two lightweight branch layers $f_\texttt{pre}$ and $f_\texttt{meta}$, followed by activation functions $a_\texttt{joint}$, $a_\texttt{pre}$, $a_\texttt{meta}$.
	The Omni-Loss consists of three losses respectively for {joint-training} $\mathcal{L}_\texttt{joint}$, {pre-training} $\mathcal{J}_\texttt{pre}$, and {meta-training} $\mathcal{J}_\texttt{meta}$, computed on the corresponding head. We also propose a {self-distillation strategy} for training the pre-flow and meta-flow, which transfers knowledge throughout the training process.}
	\label{fig:architecture}
\end{figure*}

\subsection{Omni-Net}\label{sec:omni_layer}
{Omni-Net} is a tri-flow architecture that is constructed by stacking Omni-Layers for representation learning and Omni-Heads for output prediction, as shown in Figure~\ref{fig:architecture}.

\myparagraph{Omni-Layer} We aim to simultaneously preserve the domain transferability of pre-training and the task transferability of meta-training, and promote knowledge communication between them. Thus, as shown in Figure~\ref{fig:architecture}, we design an Omni-Layer consisting of a main chunk layer $f_\texttt{joint}$ and two parallel branch layers $f_\texttt{pre}$ and $f_\texttt{meta}$. It enables three interdependent data flows with different network parameters. In the \emph{joint-flow}, the training data only go through $f_\texttt{joint}$, which is jointly trained by pre-training and meta-training to extract common knowledge as well as to coordinate the two parallel flows for a better communication between them. Besides, the two parallel data flows for pre-training and meta-training are respectively responsible for maintaining domain transferability and task transferability. For pre-training, the data pass through both $f_\texttt{joint}$ and $f_\texttt{pre}$, and then these two outputs are added as the output of this Omni-Layer in the data flow. We denote this data flow as \emph{pre-flow}. Similarly, for meta-training and its corresponding \emph{meta-flow}, the output is derived by adding the outputs of $f_\texttt{joint}$ and $f_\texttt{meta}$. Overall, the transformation function of the three parallel data flows in the $l$-th Omni-Layer can be summarized as:
\begin{equation}
	\mathbf{x}^{l}=\left\{
	\begin{array}{lcl}
		f_\texttt{joint}^{l}(\mathbf{x}^{l-1}) + f_\texttt{pre}^{l}(\mathbf{x}^{l-1})  &   & {\text{$\mathbf{x} \in $ pre-flow}}   \\
		                                                                               &   &                                         \\
		f_\texttt{joint}^{l}(\mathbf{x}^{l-1})                                         &   & {\text{$\mathbf{x} \in$ joint-flow}} \\
		                                                                               &   &                                         \\
		f_\texttt{joint}^{l}(\mathbf{x}^{l-1}) + f_\texttt{meta}^{l}(\mathbf{x}^{l-1}) &   & {\text{$\mathbf{x} \in$ meta-flow}}
	\end{array} \right.
\end{equation}

This architecture can be transformed from the layers in existing backbones by copying their original layers as the chunk layer $f_\texttt{joint}$ and adding two similar branch layers $f_\texttt{pre}$ and $f_\texttt{meta}$. We design the two parallel branches as \emph{lightweight} layers compared to the chunk layer, which maintains parameter efficiency of the Omni-Training framework. For example, if $f_\texttt{joint}$ is a convolution layer with large kernels such as $7\times7$ or $3\times3$, $f_\texttt{pre}$ and $f_\texttt{meta}$ can be convolution layers with smaller kernels such as $1\times1$. Some existing architectures may introduce some additional special layers such as batch normalization and various activation functions. We let each data flow have its specific copy of these additional layers (denoted as $a_\texttt{joint}$, $a_\texttt{pre}$ and $a_\texttt{meta}$), which strengthens the specificity of the three data flows. We omit these additional layers in the equations for simplicity.

We stack the Omni-Layers to construct the backbone for Omni-Training, and the tri-flow in each layer expands to the entire data flows in the whole backbone. Specifically, we use $F_\texttt{joint}$ to denote the overall function of the \emph{joint-flow} which stacks $f_\texttt{joint}^{l}$ in the backbone:
\begin{equation}
	F_\texttt{joint} = f_\texttt{joint}^{L} \circ \cdots \circ f_\texttt{joint}^{l} \circ \cdots \circ f_\texttt{joint}^{1}.
\end{equation}
We use $F_\texttt{pre}$ to denote the overall function of the stacked layers in the backbone that encodes the \emph{pre-flow}, which enables knowledge routing by adding the joint-flow:
\begin{equation}
	F_\texttt{pre} = \big(f_\texttt{pre}^{L} + f_\texttt{joint}^{L}\big) \circ \cdots \circ \big(f_\texttt{pre}^{l} + f_\texttt{joint}^{l}\big) \circ \cdots \circ \big(f_\texttt{pre}^{1} + f_\texttt{joint}^{1}\big).
\end{equation}
Similarly, we use $F_\texttt{meta}$ to denote the overall function of the stacked layers in the backbone that encodes the \emph{meta-flow}, which enables knowledge routing by adding the joint-flow:
\begin{equation}
	\scriptsize
	F_\texttt{meta} = \big(f_\texttt{meta}^{L} + f_\texttt{joint}^{L}\big) \circ \cdots \circ \big(f_\texttt{meta}^{l} + f_\texttt{joint}^{l}\big) \circ \cdots \circ \big(f_\texttt{meta}^{1} + f_\texttt{joint}^{1}\big).
\end{equation}
Such a stacked tri-flow encoding backbone has several benefits. First, it is parameter efficient, where the main chunk parameters are reused to encode different data flows and the architecture requires much fewer parameters than encoding these flows separately. Second, knowledge is softly shared between pre-training, meta-training, and joint-training by routing through the shared parameters in the architecture. Third, the Omni-Layer does not restrict on any specific architecture choices, but is generally applicable to various backbones in representation learning methods.

\myparagraph{Omni-Head}
The Omni-Head $H$ generates the final predictions of the three data flows with the backbone representations. Specifically, $H$ consists of three heads: a {joint-head} $H_\texttt{joint}$, a {pre-head} $H_\texttt{pre}$ and a {meta-head} $H_\texttt{meta}$. Each head takes the corresponding data flow representations in the backbone as its input and outputs the prediction. Architectures of the three heads rely on the task, \emph{e.g.}, for classification problem, the heads can be classifiers with a single fully-connected layer. The separate outputs for the three data flows enable the use of different losses to train the three flows as introduced in Omni-Loss below. By chaining the backbone and the Omni-Head, we obtain the Omni-Net architecture.

\subsection{Omni-Loss}

Based on the Omni-Net architecture, our general idea is to train the parameters of each data flow with the corresponding pre-training or meta-training algorithm, and enhance the transferability of each flow through the Omni-Loss.

\myparagraph{Joint-Training} Joint-training is performed on the joint-flow with the losses of both pre-training and meta-training. In each iteration, we sample a standard \emph{mini-batch} $\mathcal{B}$ and a \emph{task episode} $\{\mathcal{S}, \mathcal{Q}\}$ from the large-scale training set $\mathcal{D}_\texttt{train}$.
We add the pre-training loss with the mini-batch data and the meta-training loss with the sampled task on the joint-head $H_\texttt{joint}$. The joint-training loss is
\begin{equation}
	\small
	\begin{aligned}
		\label{eqn:Lcommon}
		\mathcal{L}_\texttt{joint}  &= \   \mathbb{E}_{\mathcal{B} \sim \mathcal{P}(\mathbf{x}, \mathbf{y})} \mathbb{E}_{(\mathbf{x}, \mathbf{y}) \in \mathcal{B}} \ \ell_\texttt{pre}\big(\mathbf{y}, H_\texttt{joint} \circ F_\texttt{joint}(\mathbf{x} ) \big)                     \\
		& + \                               \mathbb{E}_{(\mathcal{S}, \mathcal{Q}) \sim \mathcal{P}(\mathcal{T} )} \mathbb{E}_{(\mathbf{x}, \mathbf{y}) \in \mathcal{Q}} \ \ell_\texttt{meta}\big(\mathbf{y}, H_\texttt{joint} \circ F_\texttt{joint}(\mathbf{x} | \mathcal{S}) \big),
	\end{aligned}
\end{equation}
where $\ell_\texttt{pre}$ and $\ell_\texttt{meta}$ are the losses of pre-training and meta-training algorithms respectively. Though the joint-training extracts shared features between the two training paradigms, such a naive combination fails to endow representations with both domain transferability and task transferability simultaneously, as we have shown in Section~\ref{sec:pre or meta}. Therefore, we further perform pre-training and meta-training on the two parallel data flows respectively to explicitly preserve domain transferability and task transferability.

\myparagraph{Pre-Training} To specifically acquire domain transferability in the network, we perform pre-training on the pre-flow.
In each iteration, we feed each sample $(\mathbf{x}, \mathbf{y})$ from the mini-batch $\mathcal{B}$ into the pre-flow of the Omni-Net, going through $F_\texttt{pre}$ and $H_\texttt{pre}$, and control the final output by the pre-training loss on the pre-flow:
\begin{equation}\label{eqn:Lfeature}
	\mathcal{L}_\texttt{pre} = \mathbb{E}_{\mathcal{B} \sim \mathcal{P}(\mathbf{x}, \mathbf{y})} \mathbb{E}_{(\mathbf{x}, \mathbf{y}) \in \mathcal{B}} \ \ell_\texttt{pre}\big(\mathbf{y}, H_\texttt{pre} \circ F_\texttt{pre}(\mathbf{x} ) \big).
\end{equation}
In addition to the knowledge transfer across different branches, 
we further enhance the specific transferability on each parallel branch throughout the learning process. In order to realize it, we employ a self-distillation strategy. Let $\theta$ denote all the parameters in the backbone $F$ and the Omni-Head $H$, $i$ denote the training steps, we keep the temporal ensemble of the network during the learning process, \emph{i.e.}, an exponential moving average (EMA) of the model parameters $\widetilde{\theta}$, which is updated smoothly during training:
\begin{equation}
	\label{eqn:Meanteacher}
	\widetilde{\theta}_{i}=\alpha\widetilde{\theta}_{i-1} + (1 - \alpha)\theta_{i}.
\end{equation}
The EMA model gathers knowledge from different training stages and serves as a teacher to guide the training of the current Omni-Net. In each iteration, the EMA model transfers knowledge to each parallel branch through knowledge distillation.
We implement this idea into \emph{self-distillation regularization} for the pre-flow:
\begin{equation}\label{eqn:Rtrans_pre}
	% \begin{aligned}
		\mathcal{R}_\texttt{pre} = \mathbb{E}_{\mathcal{B} \sim \mathcal{P}(\mathbf{x}, \mathbf{y})} \mathbb{E}_{(\mathbf{x}, \mathbf{y}) \in \mathcal{B}} \   \ell_2\big(\widetilde{H}_\texttt{pre} \circ \widetilde{F}_\texttt{pre}(\mathbf{x}),  H_\texttt{pre} \circ F_\texttt{pre}(\mathbf{x}) \big),
	% \end{aligned}
\end{equation}
where $\widetilde{F}_\texttt{pre}$ and $\widetilde{H}_\texttt{pre}$ denote the mapping functions of pre-flow and pre-head in the EMA model with the temporal ensemble parameters of $\widetilde{\theta}$, and $\ell_2$ is the squared loss. The pre-training loss improved by the self-distillation for the pre-flow is
\begin{equation}
	\label{eqn:Lpre}
	\mathcal{J}_\texttt{pre} = \mathcal{L}_\texttt{pre} + \lambda \mathcal{R}_\texttt{pre},
\end{equation}
with $\lambda$ being a hyper-parameter to trade-off the original pre-training loss and the self-distillation regularization.

\myparagraph{Meta-Training} To acquire task transferability in the network, in each iteration, we perform meta-training on the meta-flow with the sampled task episode $(\mathcal{S},\mathcal{Q})$.
Data in the support set $\mathcal{S}$ are fed into the meta-flow to obtain the conditioned model. Then, each sample $(\mathbf{x}, \mathbf{y})$ from the query set $\mathcal{Q}$ passes through the meta-flow conditioned on the support set to derive the meta-training loss:
\begin{equation}\label{eqn:Lmetafeature}
	\mathcal{L}_\texttt{meta} = \mathbb{E}_{(\mathcal{S}, \mathcal{Q}) \sim \mathcal{P}(\mathcal{T} )} \mathbb{E}_{(\mathbf{x}, \mathbf{y}) \in \mathcal{Q}} \ \ell_\texttt{meta}\big(\mathbf{y}, H_\texttt{meta} \circ F_\texttt{meta}(\mathbf{x} | \mathcal{S}) \big).
\end{equation}
Similar to the pre-flow, we impose the \emph{self-distillation regularization} to improve the transferability of the meta-learned representations across the training process for the meta-flow:
\begin{equation}\label{eqn:Rtrans_meta}
	\begin{aligned}
		\mathcal{R}_\texttt{meta} = \mathbb{E}_{(\mathcal{S}, \mathcal{Q}) \sim \mathcal{P}(\mathcal{T} )} \mathbb{E}_{(\mathbf{x}, \mathbf{y}) \in \mathcal{Q}} \   \ell_2\big(\widetilde{H}_\texttt{meta} \circ \widetilde{F}_\texttt{meta}(\mathbf{x} | \mathcal{S}),  \\ H_\texttt{meta} \circ F_\texttt{meta}(\mathbf{x} | \mathcal{S})\big) ,
	\end{aligned}
\end{equation}
where $\widetilde{F}_\texttt{meta}$ and $\widetilde{H}_\texttt{meta}$ denote the mapping functions of the meta-flow and meta-head in the EMA model, and $\ell_2$ is the squared loss. The training loss for the meta-flow includes the original meta-training loss and the self-distillation regularization as
\begin{equation}
	\label{eqn:Lmeta}
	\mathcal{J}_\texttt{meta} = \mathcal{L}_\texttt{meta} + \lambda \mathcal{R}_\texttt{meta},
\end{equation}
with $\lambda$ to trade-off the original meta-training loss and the regularization term.

\subsection{Overall Framework}

\myparagraph{Training} We train Omni-Net with the Omni-Loss to perform joint-training, pre-training and meta-training simultaneously:
\begin{equation}
	\label{eqn:Ltotal}
	\mathcal{O}_\texttt{Omni} = \mathcal{J}_\texttt{pre} + \mathcal{J}_\texttt{meta} + \mathcal{L}_\texttt{joint}.
\end{equation}
With the cooperation of Omni-Net and Omni-Loss, our framework trains the two parallel flows to obtain both domain transferability and task transferability and coordinates the two parallel flows to enable their knowledge communication, addressing both challenges of \emph{domain shift} and \emph{task shift} in few-shot learning problems.

\myparagraph{Inference} During the test time, we transfer knowledge learned from Omni-Training by reusing or fine-tuning the learned model and retraining a new Omni-Head for the new tasks on the labeled data in the support set $\mathcal{S}_\texttt{test}$. Since we focus on the representation learning stage but do not focus on the test time adaptation techniques, we train the new Omni-Head consisting of a new joint-head $H^\texttt{new}_\texttt{joint}$, a new pre-head $H^\texttt{new}_\texttt{pre}$ and a new meta-head $H^\texttt{new}_\texttt{meta}$ following the corresponding algorithms we have used for pre-training and meta-training. Then for each test sample $\mathbf{x}\in \mathcal{Q}_\texttt{test}$, we predict $\mathbf{x}$ using one of the three heads or their ensemble based on the real application constraints. For example, if we need to deploy the model to a real-time prediction application, we only use the prediction of the meta-head for fast adaptation using only a few gradient updates. If there is no resource restriction, we can use the ensemble of all three heads for more accurate predictions.

\section{Omni-Training Algorithms}

We provide instantiations and implementations of the Omni-Training framework by incorporating some mainstream pre-training and meta-training algorithms. The framework can generalize to a wider variety of algorithms as shown in our experiments.

\subsection{Pre-Training Algorithms}

\myparagraph{Classification} The pre-training algorithm for classification is known as {Baseline}~\cite{cite:ICLR2019ACloserLook} in few-shot learning literature. To instantiate, $H_\texttt{pre}$ is a fully-connected layer with weights $[\mathbf{w}_1,...,\mathbf{w}_K]$ and biases $[b_1,...,b_K]$ for $K$ classes, $F_\texttt{pre}$ and  $H_\texttt{pre}$ are pre-trained on training dataset $\mathcal{D}_\texttt{train}$ by using cross-entropy as $\ell_\texttt{pre}$:
\begin{equation}
	\ell_\texttt{pre} \big(\mathbf{y}, H_\texttt{pre} \circ F_\texttt{pre}(\mathbf{x}) \big) = -\log\left(\frac{\exp(\mathbf{w}_y^T F_\texttt{pre}(\mathbf{x}))}{\sum_k \exp(\mathbf{w}_k^T F_\texttt{pre}(\mathbf{x}))}\right),
\end{equation}
where $y$ is the class index of the ground-truth class label $\mathbf{y}$ for $\mathbf{x}$.
The model is then fine-tuned on the support set $\mathcal{S}_\texttt{test}$ for the new task with a new classification head $H^\texttt{new}_\texttt{pre}$.

\myparagraph{Regression} In the pre-training algorithm for regression, we use a fully-connected layer as the pre-head $H_\texttt{pre}$ to predict the output. Here the loss is defined as the squared error between the target value $y$ and the prediction, also known as the L2 loss:
\begin{equation}
	\ell_\texttt{pre} \big(y, H_\texttt{pre} \circ F_\texttt{pre}(\mathbf{x}) \big) = \big( H_\texttt{pre} \circ F_\texttt{pre}(\mathbf{x}) - y \big) ^{2}.
\end{equation}

\myparagraph{Reinforcement Learning} In the pre-training algorithm for reinforcement learning, we use the policy gradient in REINFORCE~\cite{sutton2000policy}. The Omni-Net serves as the parameterized policy $\pi=H_\texttt{pre} \circ F_\texttt{pre}$ with a fully-connected head $H_\texttt{pre}$ to predict the action given a state. Here the loss is defined as the expected return over the policy: $\ell_\texttt{pre} \left(H_\texttt{pre} \circ F_\texttt{pre} \right)=\mathbb{E}_{\tau \sim \pi}\big[ \sum\nolimits_{t=0}^{\infty} {r}(s_t, a_t) \big]$. The gradient of the pre-training loss $\ell_\texttt{pre}$ with respect to the parameters $\theta$ of the policy $\pi$, \emph{i.e.}, the policy gradient, is defined as
\begin{equation}
	\nabla_\theta \ell_\texttt{pre} \left(H_\texttt{pre} \circ F_\texttt{pre} \right) = \sum\limits_{s} p^\pi(s) \sum\limits_{a}\nabla_\theta \pi(a|s)Q^\pi(s,a).
\end{equation}
$p^\pi(s)$ is discounted weighting of the probability of encountering states $s$ from the initial states and $Q^\pi$ is the Q-function for $\pi$ \cite{Sutton2018}.

\subsection{Meta-Training Algorithms}

\myparagraph{Model-Agnostic Meta-Learning (MAML)} In meta-training, we first consider model-agnostic meta-learning (MAML)~\cite{cite:ICML2017MAML}, a gradient-based learning rule to rapidly adapt to new tasks with few data and gradient steps. In each iteration, we sample an episode of a support set $\mathcal{S}$ and a query set $\mathcal{Q}$, and optimize the MAML loss:
\begin{equation}
	\begin{aligned}
		  & \ell_\texttt{meta}\big(\mathbf{y}, H_\texttt{meta} \circ F_\texttt{meta}(\mathbf{x} | \mathcal{S}) \big) = \ell\big(\mathbf{y}, H_\texttt{meta} \circ F_\texttt{meta}(\mathbf{x}; \theta') \big),
	\end{aligned}
\end{equation}
for each sample $(\mathbf{x},\mathbf{y}) \in \mathcal{Q}$ in the query set. Here $\theta$ is the parameters of $H_\texttt{meta}$ and $F_\texttt{meta}$ in the meta-flow, and $\theta'=\theta-\nabla_\theta\mathbb{E}_{(\mathbf{x},\mathbf{y})\in \mathcal{S}} \ \ell\big(\mathbf{y}, H_\texttt{meta} \circ F_\texttt{meta}(\mathbf{x}; \theta) \big)$ is the model parameters after a single gradient update on the support set $\mathcal{S}$. MAML has few restrictions on the model architecture and learning task, and can be widely used on various tasks such as regression, classification and reinforcement learning, by specifying the task-aware loss $\ell$.

\myparagraph{Prototypical Networks} In the few-shot learning literature, one of the well-established meta-training algorithms is {ProtoNet}~\cite{cite:NIPS2017ProtoNet}. Let $\mathcal{S}_k$ denote the samples with the class index $k$ in a support set $\mathcal{S}$ in the episode, the prototype of this class $\mathbf{c}_k$ is the mean of the embedded data in $\mathcal{S}_k$: $\mathbf{c}_k = \mathbb{E}_{(\mathbf{x}, \mathbf{y})\in{\mathcal{S}_k}}F_\texttt{meta}(\mathbf{x}) $. A metric-based classifier predicts the probability distribution of each query point $\mathbf{x}$ based on its Euclidean distances $d$ to the prototypes, which is penalized by a cross-entropy loss for classification:
\begin{equation}\label{eqn:TMProtoNet}
	\scriptsize
	\ell_\texttt{meta}\big(\mathbf{y}, H_\texttt{meta} \circ F_\texttt{meta}(\mathbf{x} | \mathcal{S}) \big) = -\log\left(\frac{\exp(-d(F_\texttt{meta}(\mathbf{x}), \mathbf{c}_y))} {\sum_{k=1}^{K}\exp(-d(F_\texttt{meta}(\mathbf{x}), \mathbf{c}_k))}\right).
\end{equation}
For new tasks, the labeled data in the support set $\mathcal{S}_\texttt{test}$ are used to compute the prototypes of each new class. Then we can classify new samples in the query set $\mathcal{Q}_\texttt{test}$ by their nearest prototype.

\section{Experiments}\label{sec:exp}
We evaluate our Omni-Training framework with comprehensive experiments on cross-task and cross-domain settings in classification, regression and reinforcement learning problems to testify the few-shot learning performances. All the codes and datasets will be available online at {\hyperlink{https://github.com/thuml/Omni-Training}{https://github.com/thuml/Omni-Training}}.

\subsection{Classification}

\begin{table*}[tbp]
	\begin{center}
		\caption{The results of the new tasks with $5$ or $1$ labeled samples per class on mini-ImageNet and CUB datasets.}% The backbone used is consistent with original papers.which forms the widely-evaluated \emph{cross-task} setting}
		\label{table:Accuracy1}
		%\vskip 0.05in
		%\begin{large}
		%\begin{sc}
		\addtolength{\tabcolsep}{5pt}
		% \resizebox{0.9\textwidth}{!}{%
			\begin{tabular}{lccccc}
				\toprule
				% 	\hline
				\multirow{2}{40pt}{\textbf{Method}} &\multirow{2}{40pt}{\textbf{Backbone}}&\multicolumn{2}{c}{\textbf{mini-ImageNet}}&\multicolumn{2}{c}{\textbf{CUB}} \\
				                                                      &           & $K=5$                   & $K=1$                   & $K=5$                   & $K=1$                   \\
				\midrule
				% 	\hline
				MatchingNet~\cite{cite:NIPS2016MatchingNet}          & ResNet-18 & $68.88\pm0.69$          & $52.91\pm0.88$          & $83.64\pm0.60$          & $72.36\pm0.90$          \\
				ProtoNet~\cite{cite:NIPS2017ProtoNet}                & ResNet-18 & $73.68\pm0.65$          & $54.16\pm0.82$          & $87.42\pm0.48$          & $71.88\pm0.91$          \\
				RelationNet~\cite{cite:CVPR2018RelationNet}          & ResNet-18 & $69.83\pm0.68$          & $52.48\pm0.86$          & $82.75\pm0.58$          & $67.59\pm1.02$          \\
				MAML~\cite{cite:ICML2017MAML}                        & ResNet-18 & $65.72\pm0.77$          & $49.61\pm0.92$          & $82.70\pm0.65$          & $69.96\pm1.01$          \\
				TADAM~\cite{cite:NIPS2018TADAM}                      & ResNet-12 & $76.70\pm0.30$          & $58.50\pm0.30$          & $-$                     & $-$                     \\
				GNN~\cite{cite:ICLR2018GNN}                          & ResNet-18 & $78.80\pm0.78$          & $57.40\pm0.98$          & $90.74\pm0.57$          & $78.52\pm1.03$          \\
				LEO~\cite{cite:ICLR2019LEO}                          & WRN28-10  & $77.59\pm0.12$          & $61.76\pm0.08$          & $78.27\pm0.16$          & $68.22\pm0.22$          \\
				Baseline~\cite{cite:ICLR2019ACloserLook}             & ResNet-18 & $74.27\pm0.63$          & $51.75\pm0.80$          & $82.85\pm0.55$          & $65.51\pm0.87$          \\
				Baseline++~\cite{cite:ICLR2019ACloserLook}           & ResNet-18 & $75.68\pm0.63$          & $51.87\pm0.77$          & $83.58\pm0.54$          & $67.02\pm0.90$          \\
				MTL~\cite{cite:CVPR2019MetaTransferLearning}         & ResNet-12 & $75.50\pm0.80$          & $61.20\pm1.80$          & $-$                     & $-$                     \\
				MetaOpt~\cite{cite:CVPR2019MetaOpt}                  & ResNet-12 & $78.63\pm0.36$          & $62.64\pm0.61$          & $90.90\pm0.23$          & $80.23\pm0.44$          \\
				TapNet~\cite{cite:ICML2019TapNet}                    & ResNet-12 & $76.36\pm0.10$          & $61.65\pm0.15$          & $-$                     & $-$                     \\
				Robust20~\cite{cite:ICCV2019EnsembleFewShot}         & ResNet-18 & $81.59\pm0.42$          & $63.95\pm0.61$          & $84.62\pm0.44$          & $69.47\pm0.69$          \\
				CAN~\cite{cite:NIPS2019CrossAttention}               & ResNet-12 & $79.44\pm0.34$          & $63.85\pm0.48$          & $-$                     & $-$                     \\
				RFS~\cite{cite:ECCV2020RethinkingFewShot}            & ResNet-12 & $82.14\pm0.43$          & $64.82\pm0.60$          & $-$                     & $-$                     \\
				Neg-Margin~\cite{cite:ECCV2020NegativeMargin}        & ResNet-18 & $80.94\pm0.59$          & $62.33\pm0.82$          & $89.40\pm0.40$          & $72.66\pm0.90$          \\
				PMM~\cite{cite:Arxiv2020PMM}                         & ResNet-18 & $77.76\pm0.58$          & $60.11\pm0.73$          & $86.01\pm0.50$          & $73.94\pm1.10$          \\
				Multi-Task~\cite{cite:ICML2021MultiTaskMetaLearning} & ResNet-12 & $77.72\pm0.09$          & $59.84\pm0.22$          & $-$                     & $-$                     \\
				Meta-Maxup~\cite{cite:ICML2021DataAugMetaLearning}   & ResNet-12 & $79.38\pm0.24$          & $62.81\pm0.34$          & $-$                     & $-$                     \\
				\midrule
				% 	\hline
				OT-Proto                             & ResNet-18 & $81.26\pm0.57$          & $64.31\pm0.86$          & $91.09\pm0.38$          & $81.18\pm0.78$          \\
				OT-Proto                             & ResNet-12 & ${82.36\pm0.54}$        & ${66.62\pm0.80}$        & ${91.93\pm0.38}$        & ${82.94\pm0.73}$        \\
				OT-GNN                               & ResNet-18 & $\mathbf{87.14\pm0.59}$ & $\mathbf{70.99\pm0.97}$ & $\mathbf{95.96\pm0.33}$ & $\mathbf{87.73\pm0.78}$ \\
				\bottomrule
				% 	\hline
			\end{tabular}%
		% }
		%\end{sc}
		%{large}
	\end{center}
	%\vskip -0.1in
\end{table*}

\myparagraph{Datasets} We consider few-shot classification problems with four datasets: in-domain datasets \textbf{mini-ImageNet}~\cite{cite:NIPS2016MatchingNet}, \textbf{CUB}~\cite{cite:CUB}, and cross-domain datasets \textbf{mini-ImageNet$\rightarrow$CUB}, \textbf{Multi-domain}. \textbf{mini-ImageNet} is a subset of ILSVRC-12 dataset~\cite{cite:IJCV2015ImageNet} for generic object recognition. It contains $100$ classes with $600$ images per class. We use the same split introduced by~\cite{cite:ICLR2017OptimizationasaModel}, which respectively splits $64$/$16$/$20$ classes for the training/validation/testing set. \textbf{CUB} is a fine-grained dataset of birds with a total of $200$ classes and $11,788$ images. We follow the protocol of~\cite{cite:Arxiv2018MetricAgnostic} and split the dataset into $100$/$50$/$50$ classes for training/validation/testing. \textbf{mini-ImageNet$\rightarrow$CUB} is a cross-domain dataset. Following~\cite{cite:ICLR2019ACloserLook}, we use mini-ImageNet dataset as the training set and split the CUB set as $50$/$50$ classes for validation and testing. \textbf{Multi-domain} is another cross-domain dataset. We follow the split in~\cite{cite:ICLR2020CrossDomain} and use the datasets of mini-ImageNet, CUB, Cars~\cite{cite:ICCV2013Cars}, Places~\cite{cite:TPAMI2017Places} and Plantae~\cite{cite:CVPR2018Plantae} as different domains. We explore two settings. The first is training the model on the mini-ImageNet domain and evaluating on other four domains. The second is the leave-one-out setting which selects one domain for evaluation and trains the model with all other domains.

\begin{table}[tbp]
	\begin{center}
		\caption{The classification accuracy of the new tasks with $5$ or $1$ labeled samples per class in the \emph{cross-domain} setting, mini-ImageNet$\rightarrow$CUB.}
		\label{table:Accuracy2}
		%\vskip 0.05in
		%\begin{large}
		%\begin{sc}
		%\addtolength{\tabcolsep}{-4.5pt}
		% \resizebox{0.55\textwidth}{!}{%
			\begin{tabular}{lcc}
				\toprule
				% \hline
				\textbf{Method}                                & $K=5$                   & $K=1$                   \\
				\midrule
				% \hline
				MatchingNet \cite{cite:NIPS2016MatchingNet}           & $53.07\pm0.74$          & $38.78\pm0.73$          \\
				ProtoNet \cite{cite:NIPS2017ProtoNet}                 & $62.02\pm0.70$          & $40.07\pm0.75$          \\
				RelationNet \cite{cite:CVPR2018RelationNet}           & $57.71\pm0.73$          & $37.71\pm0.69$          \\
				MAML \cite{cite:ICML2017MAML}                         & $51.34\pm0.72$          & $40.15\pm0.65$          \\
				GNN \cite{cite:ICLR2018GNN}                           & $65.56\pm0.87$          & $43.65\pm0.86$          \\
				Baseline \cite{cite:ICLR2019ACloserLook}              & $65.57\pm0.70$          & $43.59\pm0.74$          \\
				Baseline++ \cite{cite:ICLR2019ACloserLook}            & $62.04\pm0.76$          & $44.14\pm0.77$          \\
				Robust20 \cite{cite:ICCV2019EnsembleFewShot}          & $65.04\pm0.57$          & $-$                     \\
				Neg-Margin
				\cite{cite:ECCV2020NegativeMargin}                    & $67.03\pm0.76$          & $-$                     \\
				PMM
				\cite{cite:Arxiv2020PMM}                              & $68.77\pm0.90$          & $-$                     \\
				\midrule
				% \hline
				OT-Proto                              & $71.30\pm0.71$          & $50.42\pm0.82$          \\
				OT-GNN                                & $\mathbf{75.83\pm0.82}$ & ${50.89\pm0.91}$        \\
				\bottomrule
				% \hline
			\end{tabular}%
		% }
		%\end{sc}
		%{large}
	\end{center}
	%\vskip -0.1in
\end{table}

\myparagraph{Implementation Details} We use ResNet-18 in~\cite{cite:ICLR2019ACloserLook} and ResNet-12 with dropblocks in~\cite{cite:NIPS2018TADAM} as the backbone for {mini-ImageNet}, {CUB} and {mini-ImageNet$\rightarrow$CUB}. Following~\cite{cite:ICLR2020CrossDomain}, we use ResNet-10 on Multi-domain for a fair comparison. We refactor ResNet into a backbone for Omni-Training by transforming all convolution layers into Omni-Layers, where each Omni-Layer uses the $1\times1$ convolution layer as the lightweight branch layer. We employ {Baseline} in~\cite{cite:ICLR2019ACloserLook} as the pre-training method and explore two powerful meta-training methods, {ProtoNet}~\cite{cite:NIPS2017ProtoNet} and GNN~\cite{cite:ICLR2018GNN}, denoted as {OT-Proto} and {OT-GNN} respectively. In each iteration, a mini-batch is sampled with the batch size of $64$ for pre-training, and an episode of task is sampled for meta-training, with a support set containing $5$ categories each having $5$ labeled instances, and a query set containing the same categories with $16$ instances per class. We apply standard data augmentation including random crop, left-right flip and color jitter to the training samples. We train our framework with $100$ epochs for the mini-ImageNet, mini-ImageNet$\rightarrow$CUB and Multi-domain datasets, and with $400$ epochs for the CUB dataset. We use accuracy on the validation set to choose the best model for testing. In the test stage, we randomly sample $600$ tasks from the testing set. Each task contains $5$ unseen classes with $K=5$ or $K=1$ labeled samples per class as the support set, and another $16$ instances per class as the query set to be predicted. The average accuracy as well as the $95\%$ confidence intervals are reported. The hyper-parameter is chosen as $\alpha=0.99$, $\lambda=3.0$. We train the networks from scratch and use Adam optimizer~\cite{cite:Arxiv2014Adam} with an initial learning rate of $0.001$.

\begin{table*}[t]
	\begin{center}
		\caption{The results of the tasks from unseen domains with $5$ or $1$ labeled samples per class in the \emph{Multi-domain} setting (trained with mini-ImageNet).}
		\label{table:Accuracy3}
		%\vskip 0.05in
		\resizebox{\textwidth}{!}{%
			%\begin{Huge}
			%\begin{sc}
			%\renewcommand\tabcolsep{1.5pt}
			\begin{tabular}{lcccccccc}
				\toprule
				% \hline
				\multirow{2}{40pt}{\textbf{Method}}&\multicolumn{2}{c}{\textbf{CUB}}&\multicolumn{2}{c}{\textbf{Cars}}&\multicolumn{2}{c}{\textbf{Places}}&\multicolumn{2}{c}{\textbf{Plantae}} \\
				                                                   & $K=5$                   & $K=1$                   & $K=5$                   & $K=1$                   & $K=5$                   & $K=1$                   & $K=5$                   & $K=1$                   \\
				\midrule
				% \hline
				MatchingNet~\cite{cite:NIPS2016MatchingNet} & $51.37\pm0.77$          & $35.89\pm0.51$          & $38.99\pm0.64$          & $30.77\pm0.47$          & $63.16\pm0.77$          & $49.86\pm0.79$          & $46.53\pm0.68$          & $32.70\pm0.60$          \\
				ProtoNet~\cite{cite:NIPS2017ProtoNet}       & $57.64\pm0.85$          & $38.18\pm0.76$          & $42.84\pm0.73$          & $29.72\pm0.59$          & $68.86\pm0.70$          & $49.24\pm0.81$          & $47.41\pm0.70$          & $35.02\pm0.63$          \\
				RelationNet~\cite{cite:CVPR2018RelationNet} & $57.77\pm0.69$          & $42.44\pm0.77$          & $37.33\pm0.68$          & $29.11\pm0.60$          & $63.32\pm0.76$          & $48.64\pm0.85$          & $44.00\pm0.60$          & $33.17\pm0.64$          \\
				GNN~\cite{cite:ICLR2018GNN}                 & $62.25\pm0.65$          & $45.69\pm0.68$          & $44.28\pm0.63$          & $31.79\pm0.51$          & $70.84\pm0.65$          & $53.10\pm0.80$          & $52.53\pm0.59$          & $35.60\pm0.56$          \\
				FT-Matching~\cite{cite:ICLR2020CrossDomain} & $55.23\pm0.83$          & $36.61\pm0.53$          & $41.24\pm0.65$          & $29.82\pm0.44$          & $64.55\pm0.75$          & $51.07\pm0.68$          & $41.69\pm0.63$          & $34.48\pm0.50$          \\
				FT-Relation~\cite{cite:ICLR2020CrossDomain} & $59.46\pm0.71$          & $44.07\pm0.77$          & $39.91\pm0.69$          & $28.63\pm0.59$          & $66.28\pm0.72$          & $50.68\pm0.87$          & $45.08\pm0.59$          & $33.14\pm0.62$          \\
				FT-GNN~\cite{cite:ICLR2020CrossDomain}      & ${66.98\pm0.68}$        & ${47.47\pm0.75}$        & $44.90\pm0.64$          & $31.61\pm0.53$          & $73.94\pm0.67$          & ${55.77\pm0.79}$        & $53.85\pm0.62$          & $35.95\pm0.58$          \\
				\midrule
				% \hline
				OT-Proto                          & $65.17\pm0.75$          & $45.83\pm0.78$          & $\mathbf{51.19\pm0.71}$ & ${34.82\pm0.70}$        & ${74.16\pm0.69}$        & $55.73\pm0.89$          & ${57.88\pm0.69}$        & ${39.51\pm0.71}$        \\
				OT-GNN                            & $\mathbf{70.24\pm0.82}$ & $\mathbf{49.51\pm0.96}$ & ${48.99\pm0.83}$        & $\mathbf{35.31\pm0.78}$ & $\mathbf{79.61\pm0.81}$ & $\mathbf{61.74\pm1.05}$ & $\mathbf{60.08\pm0.81}$ & $\mathbf{40.52\pm0.81}$ \\
				\bottomrule
				% \hline
			\end{tabular}%
			%\end{sc}
			%\end{Huge}
		}
	\end{center}
	%\vskip -0.21in
\end{table*}

\begin{table*}[t]
	\begin{center}
		\caption{The results of the tasks from unseen domains with $5$ or $1$ labeled samples per class in the \emph{Multi-domain} setting (trained with all other domains).}
		\label{table:Accuracy4}
		%\vskip 0.05in
		\resizebox{\textwidth}{!}{%
			%\begin{Huge}
			%\begin{sc}
			%\renewcommand\tabcolsep{1.5pt}
			\begin{tabular}{lcccccccc}
				\toprule
				% \hline
				\multirow{2}{40pt}{\textbf{Method}}&\multicolumn{2}{c}{\textbf{CUB}}&\multicolumn{2}{c}{\textbf{Cars}}&\multicolumn{2}{c}{\textbf{Places}}&\multicolumn{2}{c}{\textbf{Plantae}} \\
				                                                   & $K=5$                   & $K=1$                   & $K=5$                   & $K=1$                   & $K=5$                   & $K=1$                   & $K=5$                   & $K=1$                   \\
				\midrule
				% \hline
				MatchingNet~\cite{cite:NIPS2016MatchingNet} & $51.92\pm0.80$          & $37.90\pm0.55$          & $39.87\pm0.51$          & $28.96\pm0.45$          & $61.82\pm0.57$          & $49.01\pm0.65$          & $47.29\pm0.51$          & $33.21\pm0.51$          \\
				ProtoNet~\cite{cite:NIPS2017ProtoNet}       & $59.26\pm0.89$          & $39.31\pm0.72$          & $43.66\pm0.68$          & $29.52\pm0.54$          & $68.03\pm0.61$          & $47.96\pm0.77$          & $49.35\pm0.72$          & $35.40\pm0.68$          \\
				RelationNet~\cite{cite:CVPR2018RelationNet} & $62.13\pm0.74$          & $44.33\pm0.59$          & $40.64\pm0.54$          & $29.53\pm0.45$          & $64.34\pm0.57$          & $47.76\pm0.63$          & $46.29\pm0.56$          & $33.76\pm0.52$          \\
				GNN~\cite{cite:ICLR2018GNN}                 & $69.26\pm0.68$          & $49.46\pm0.73$          & $48.91\pm0.67$          & $32.95\pm0.56$          & $72.59\pm0.67$          & $51.39\pm0.80$          & $58.36\pm0.68$          & $37.15\pm0.60$          \\
				FT-Matching~\cite{cite:ICLR2020CrossDomain} & $61.41\pm0.57$          & $43.29\pm0.59$          & $43.08\pm0.55$          & $30.62\pm0.48$          & $64.99\pm0.59$          & $52.51\pm0.67$          & $48.32\pm0.57$          & $35.12\pm0.54$          \\
				FT-Relation~\cite{cite:ICLR2020CrossDomain} & $64.99\pm0.54$          & $48.38\pm0.63$          & $43.44\pm0.59$          & $32.21\pm0.51$          & $67.35\pm0.54$          & $50.74\pm0.66$          & $50.39\pm0.52$          & $35.00\pm0.52$          \\
				FT-GNN~\cite{cite:ICLR2020CrossDomain}      & $\mathbf{73.11\pm0.68}$ & $\mathbf{51.51\pm0.80}$ & $49.88\pm0.67$          & $34.12\pm0.63$          & $77.05\pm0.65$          & ${56.31\pm0.80}$        & $58.84\pm0.66$          & $42.09\pm0.68$          \\
				\midrule
				% \hline
				OT-Proto                          & $67.76\pm0.74$          & $46.62\pm0.77$          & ${52.02\pm0.74}$        & ${36.36\pm0.70}$        & ${73.57\pm0.66}$        & $52.20\pm0.81$          & ${59.37\pm0.69}$        & ${40.95\pm0.66}$        \\
				OT-GNN                            & $71.78\pm0.83$          & $49.78\pm0.94$          & $\mathbf{54.39\pm0.83}$ & $\mathbf{37.00\pm0.79}$ & $\mathbf{78.03\pm0.80}$ & $\mathbf{58.27\pm0.99}$ & $\mathbf{62.22\pm0.79}$ & $\mathbf{43.02\pm0.85}$ \\
				\bottomrule
				% \hline
			\end{tabular}%
			%\end{sc}
			%\end{Huge}
		}
	\end{center}
	%\vskip -0.21in
\end{table*}

\myparagraph{Results on Cross-Task Benchmarks} We first evaluate our method on the general dataset {mini-ImageNet} and the fine-grained dataset {CUB}. These two scenarios are considered as \emph{cross-task} benchmarks, as the training and testing data are from the same domain. The results with $K=5$ and $K=1$ are shown in Table~\ref{table:Accuracy1}. Omni-Training outperforms corresponding pre-training and meta-training methods, especially in extremely difficult scenarios with only $1$ labeled instance. Note that although from the same dataset, there still exists domain shift between the training and test sets caused by the split of different label sets. Our framework manages to incorporate pre-training and meta-training effectively to acquire both domain transferability and task transferability and thus achieves higher performance. Omni-Training outperforms state-of-the-art algorithms, including MTL~\cite{cite:CVPR2019MetaTransferLearning} which combines pre-training and meta-training sequentially. This confirms that our design can better bridge pre-training and meta-training.

\myparagraph{Results on Cross-Domain Benchmarks} We consider two more challenging \emph{cross-domain} benchmarks, {mini-imageNet$\rightarrow$CUB} and {Multi-domain}. Different from the cross-task benchmarks discussed above, in the cross-domain setting, the testing data are not only from different classes, but also from different domains, causing greater domain shift between the training and testing data. As shown in Table~\ref{table:Accuracy2}, meta-training algorithms degrade due to the domain shift while pre-training algorithms generalize better to the unseen domain. Omni-Training outperforms meta-training methods by a large margin, indicating the significance of domain transferability in the cross-domain setting. Also, Omni-Training outperforms the pre-training Baseline, which reveals the importance of task transferability to fully enable few-shot learning.

A more challenging benchmark is {Multi-domain} with more domains and larger domain shift. Table~\ref{table:Accuracy3} reports the results of the first setting where we train on the mini-ImageNet domain and test on other four domains. Table~\ref{table:Accuracy4} reports the results of the second leave-one-out setting, where we choose one domain as the unseen test domain and train the model with all other domains. We specially compare Omni-Training with Feature-Transformation~\cite{cite:ICLR2020CrossDomain}, which is a framework adopts domain generalization~\cite{cite:NIPS2011DG} into meta-training to obtain both domain transferability and task transferability.
Among its implementations, FT-GNN achieves the best performance by incorporating a strong meta-training algorithm, GNN~\cite{cite:ICLR2018GNN}. When trained on mini-ImageNet, Omni-Training can still achieve comparable or better performance than FT-GNN with a simple meta-training algorithm such as ProtoNet. We also incorporate GNN into Omni-Training to form OT-GNN, which generally outperforms FT-GNN in most tasks. Note that FT-GNN has a special design for domain generalization, which is tailored for the multi-domain setting. But OT-GNN also achieves better performance on most cases, confirming that Omni-Training works generally well in different situations.

\subsection{Regression}

\myparagraph{Datasets} For few-shot regression problems, we conduct experiments on a sinusoid dataset following~\cite{cite:ICML2017MAML}. Specifically, the regression problem is to predict the output $y$ on a sine wave given the input $x$. We define a task as regressing a sine wave with a particular amplitude and phase from some labeled data and consider a continuous task distribution in which the amplitude varies within $[0.1, 5.0]$ and the phase varies within $[0, 2\pi]$. The input datapoint $x$ is sampled uniformly from $[-5.0, 5.0]$ for all tasks. The training dataset $\mathcal{D}_\texttt{train}$ contains a large number of sampled sine waves and each test task $\{\mathcal{S}_\texttt{test}, \mathcal{Q}_\texttt{test}\}$ is an unseen sinusoid with a few labeled datapoints in $\mathcal{S}_\texttt{test}$ and other points which need prediction in $\mathcal{Q}_\texttt{test}$. The goal is to train a regression model on $\mathcal{D}_\texttt{train}$ to predict the outputs of the datapoints in the query set $\mathcal{Q}_\texttt{test}$ after adaptation with a few labeled data in $\mathcal{S}_\texttt{test}$.

\myparagraph{Implementation Details} We take the mean-squared error between the predictions and ground-truth values as the training loss. We use {Baseline}~\cite{cite:ICLR2019ACloserLook} for pre-training and use \texttt{MAML}~\cite{cite:ICML2017MAML} for meta-training. We employ a backbone with $2$ fully-connected layers of size $64$ with the activation function of Tanh. The training set $\mathcal{D}_\texttt{train}$ has 30000 randomly sampled tasks and each task is a sine wave with $50$ labeled datapoints. We then enable few-shot regression on a new sine wave with a support set of $K=\{5,10,20\}$ labeled examples and test the adapted model on points in $\mathcal{Q}_\texttt{test}$ of the wave. We train the model on $\mathcal{D}_\texttt{train}$ and fine-tune it on the $K$ labeled examples for the new sine wave with an SGD optimizer. The learning rate for the inner loop is $0.02$ and that for parameter update is initialized as $0.01$.

\begin{figure*}[tbp]
	\centering
	\subfloat[$K=5$]{
		\includegraphics[width=0.31\textwidth]{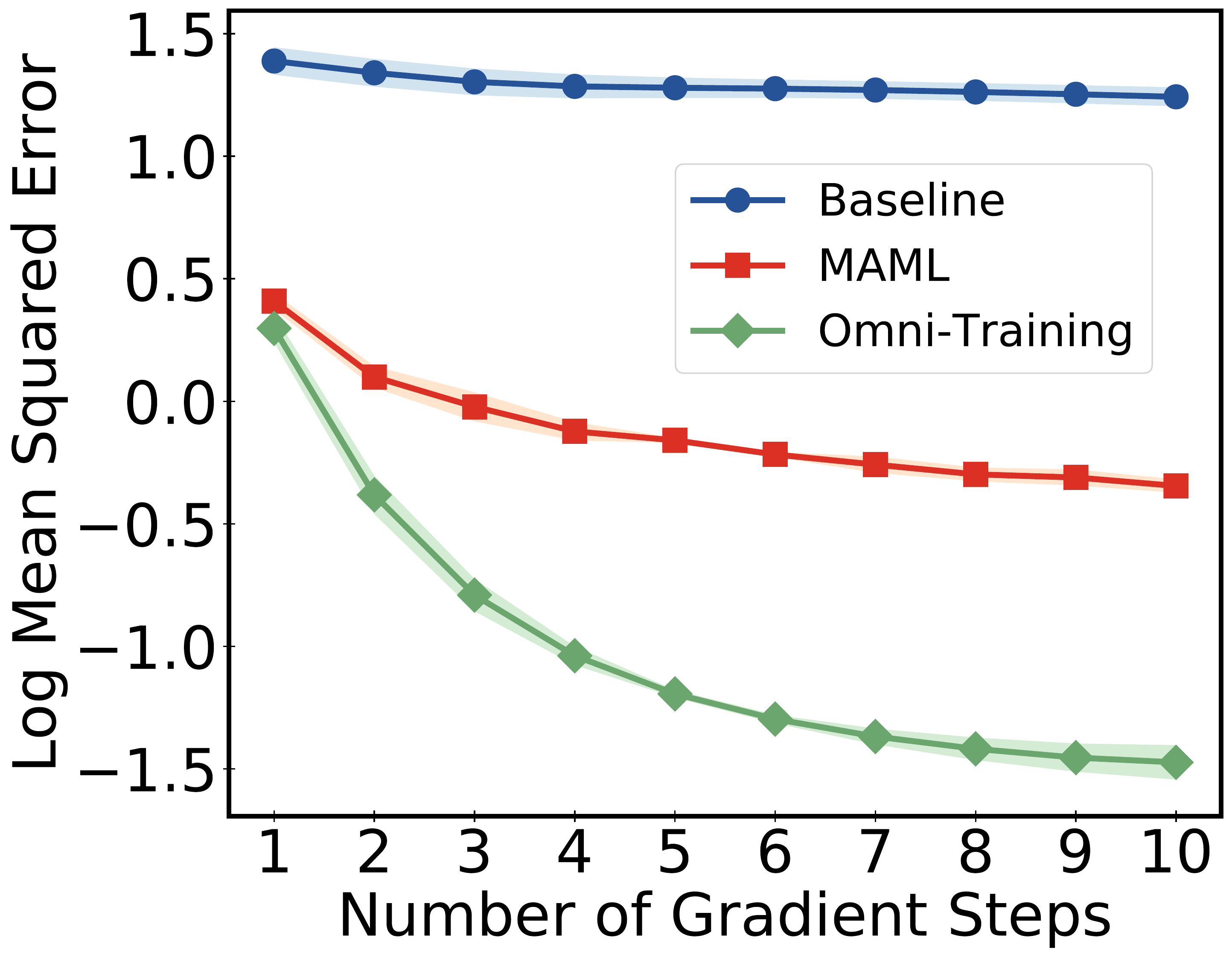} \label{fig:Regression_5shot}}
	\hfil
	\subfloat[$K=10$]{
		\includegraphics[width=0.31\textwidth]{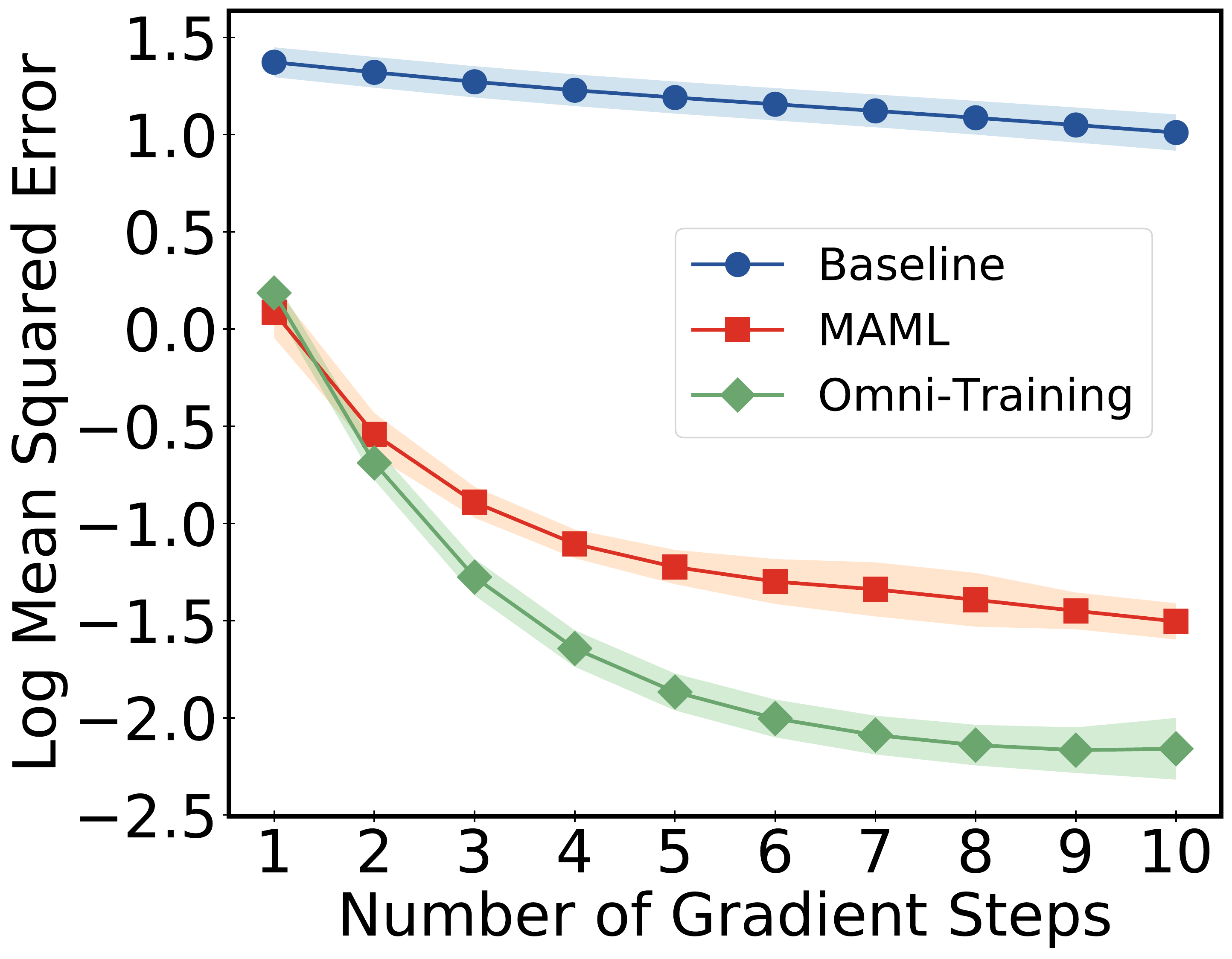} \label{fig:Regression_10shot}}
	\hfil
	\subfloat[$K=20$]{
		\includegraphics[width=0.31\textwidth]{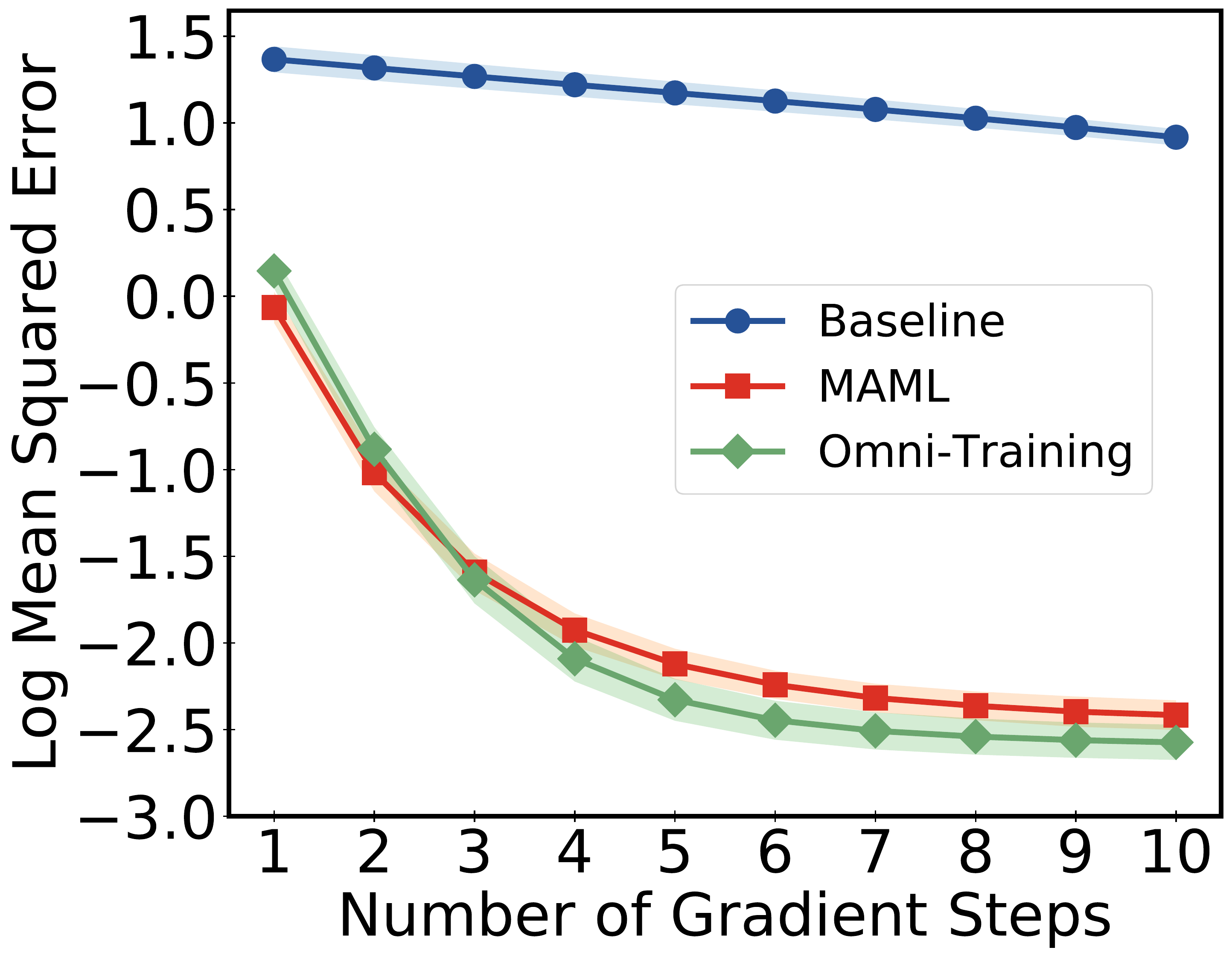} \label{fig:Regression_20shot}}
	\hfil
	\caption{The few-shot regression results of different training methods with different gradients steps and different support set sizes $K$.}
	\label{fig:Regression}
\end{figure*}

\begin{figure*}[tbp]
	\centering
	\subfloat[Baseline]{
		\includegraphics[width=0.31\textwidth]{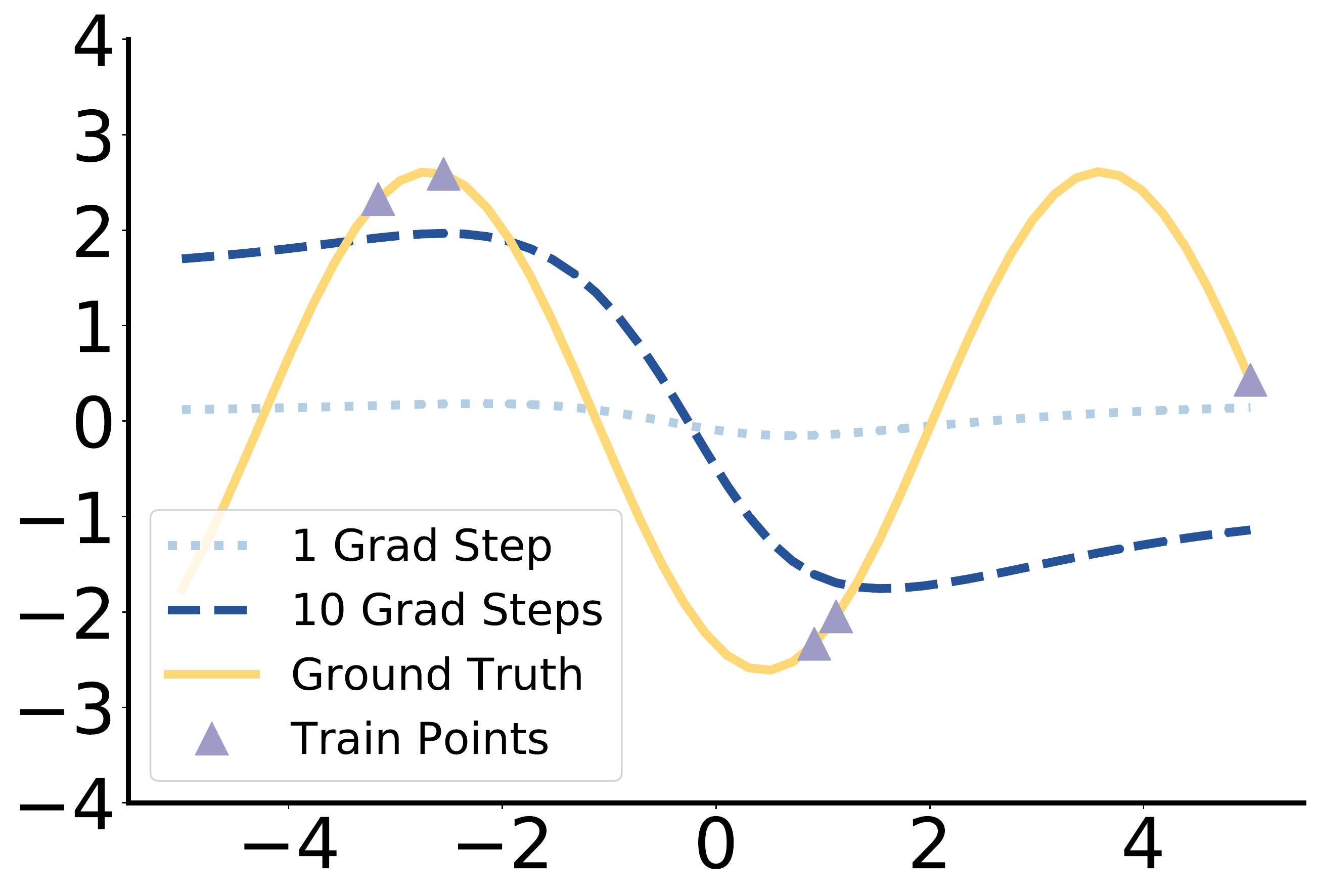} \label{fig:Regression_case1}}
	\hfil
	\subfloat[MAML]{
		\includegraphics[width=0.31\textwidth]{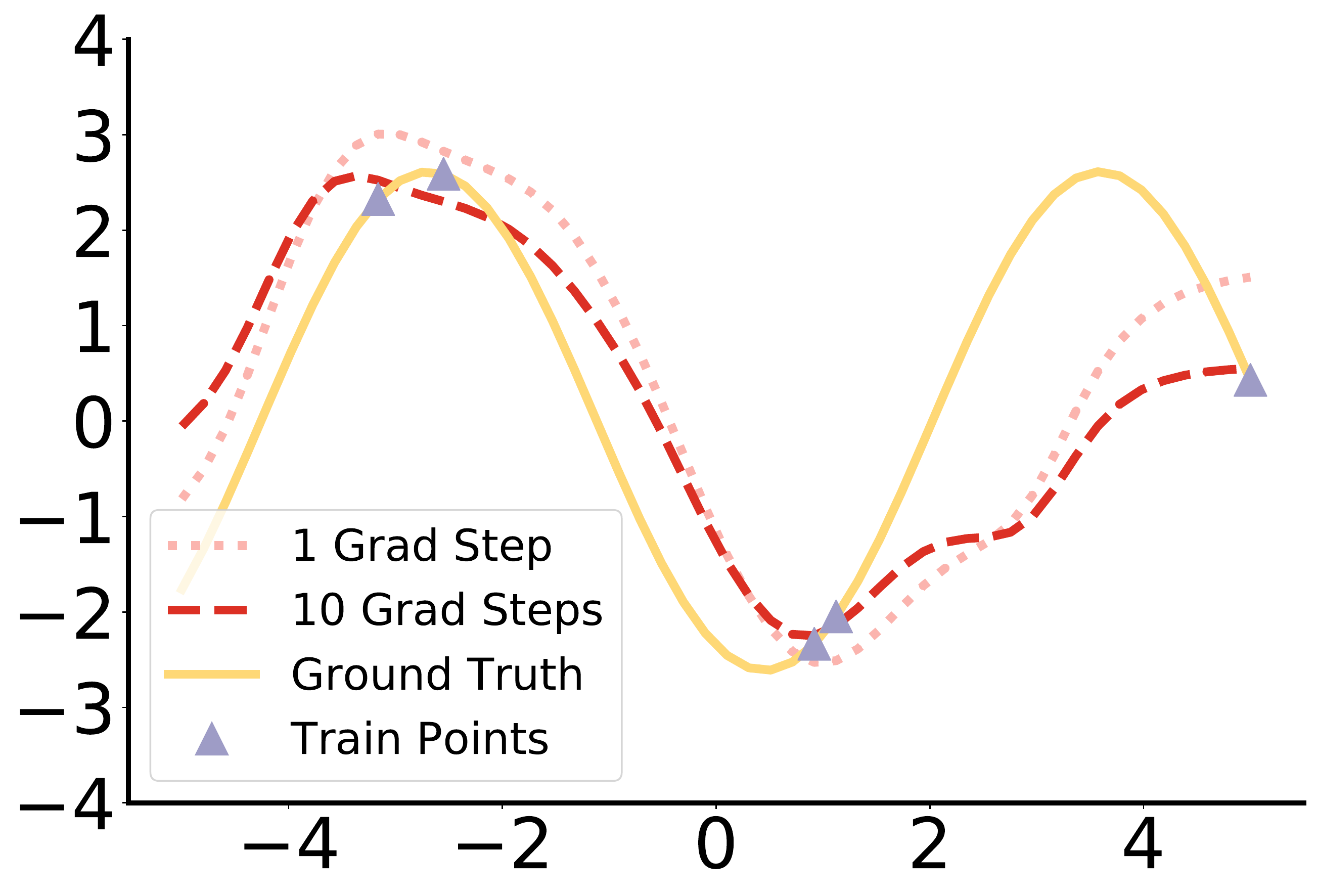} \label{fig:Regression_case2}}
	\hfil
	\subfloat[Omni-Training]{
		\includegraphics[width=0.31\textwidth]{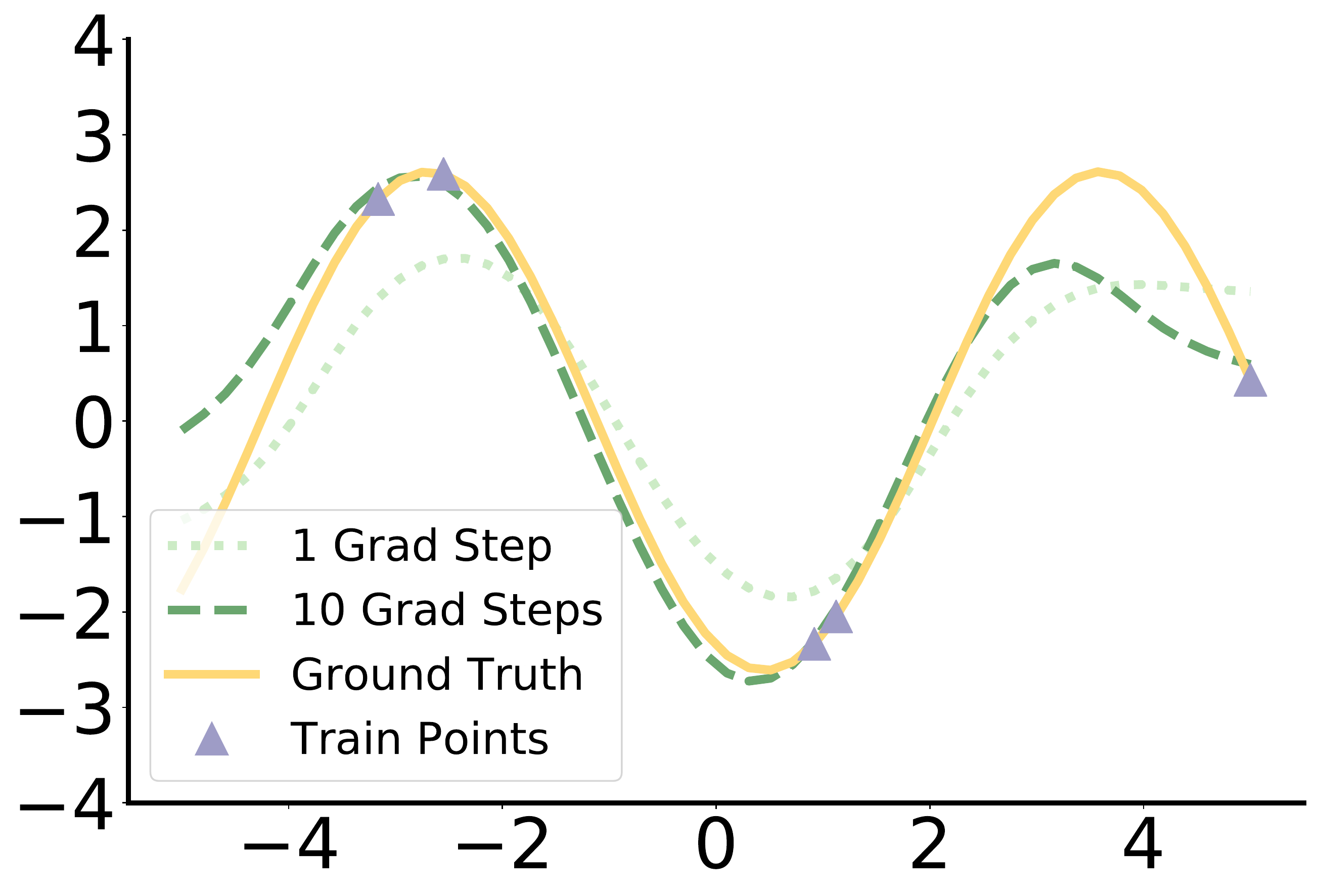} \label{fig:Regression_case3}}
	\hfil
	\caption{The recovered sine wave of Baseline, MAML and Omni-Training. The models are updated using $5$ sampled points with $1$ or $10$ gradient steps.}
	\label{fig:Regression_case}
\end{figure*}

\myparagraph{Results}
We sample 100 new tasks for testing and report the mean squared error after fine-tuning with different gradient steps from $1$ to $10$. As shown in Figure~\ref{fig:Regression}, Baseline generally performs worse than MAML. The tasks change rapidly during the training and test stages in this problem and task transferability is important, which is missing for pre-training methods. With different numbers of labeled data and of gradient steps, Omni-Training consistently improves upon the meta-training method, which shows the efficacy of Omni-Training for regression tasks.

We further conduct a case study and show the typical sine waves recovered by pre-training, meta-training and Omni-Training with $5$ labeled samples and with $1$ or $10$ gradient steps in Figure~\ref{fig:Regression_case}. We also show the ground-truth sine wave and the labeled points in the support set. MAML and Omni-Training quickly regress closed to the ground-truth curve, while the process is much slower for Baseline. Compared with MAML, the recovered curve of Baseline maintains smooth, which is an important common property in the sinusoid distribution. Omni-Training also maintains a smoother curve, which simultaneously fits these datapoints quickly and preserves the domain transferability of sine waves. This explains the improvements brought by the Omni-Training framework.

\begin{figure*}[tbp]
	\centering
	\subfloat[2D Navigation]{
		\includegraphics[width=0.31\textwidth]{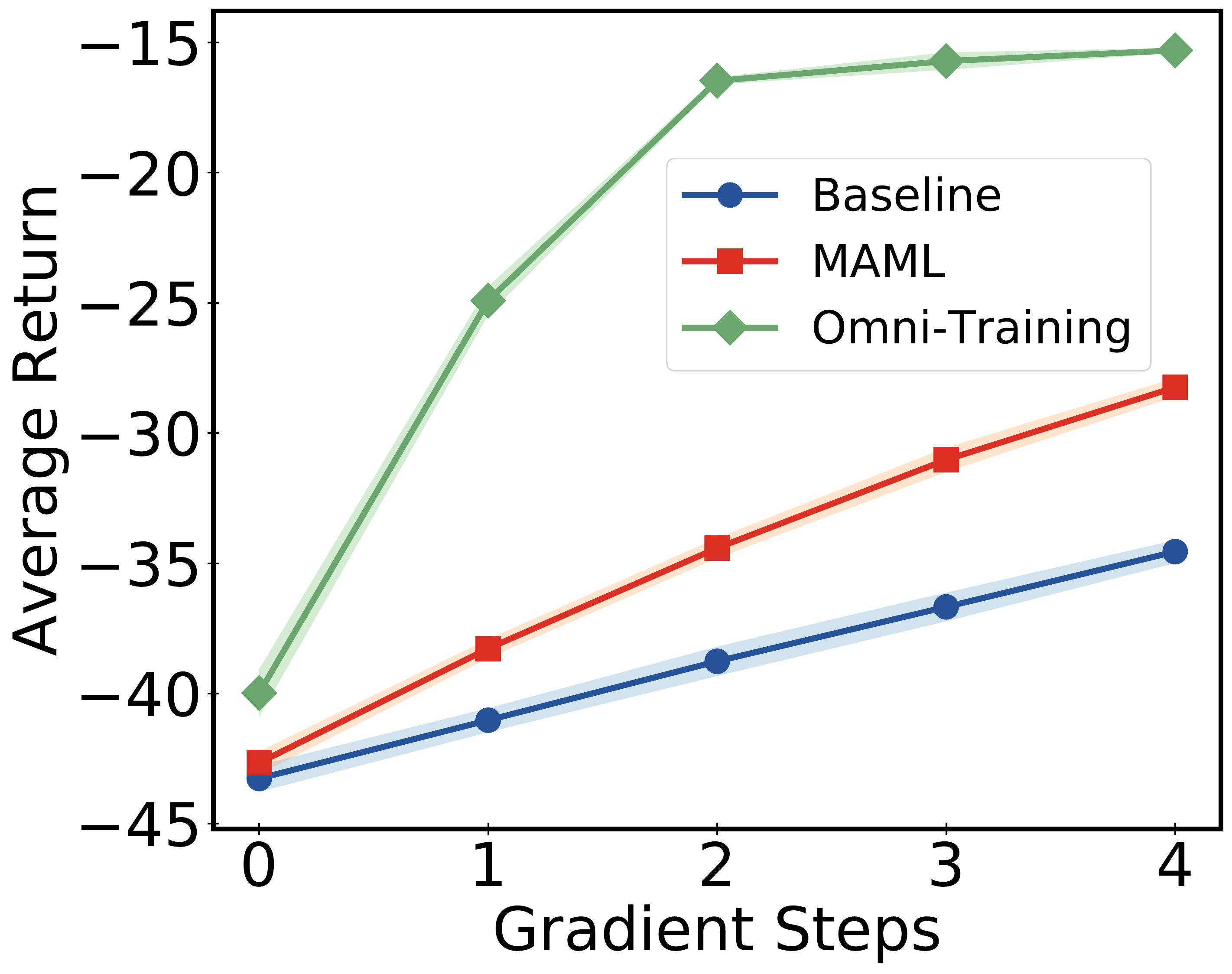} \label{fig:Reinforcement_Learning_1}}
	\hfil
	\subfloat[Locomotion-Velocity]{
		\includegraphics[width=0.31\textwidth]{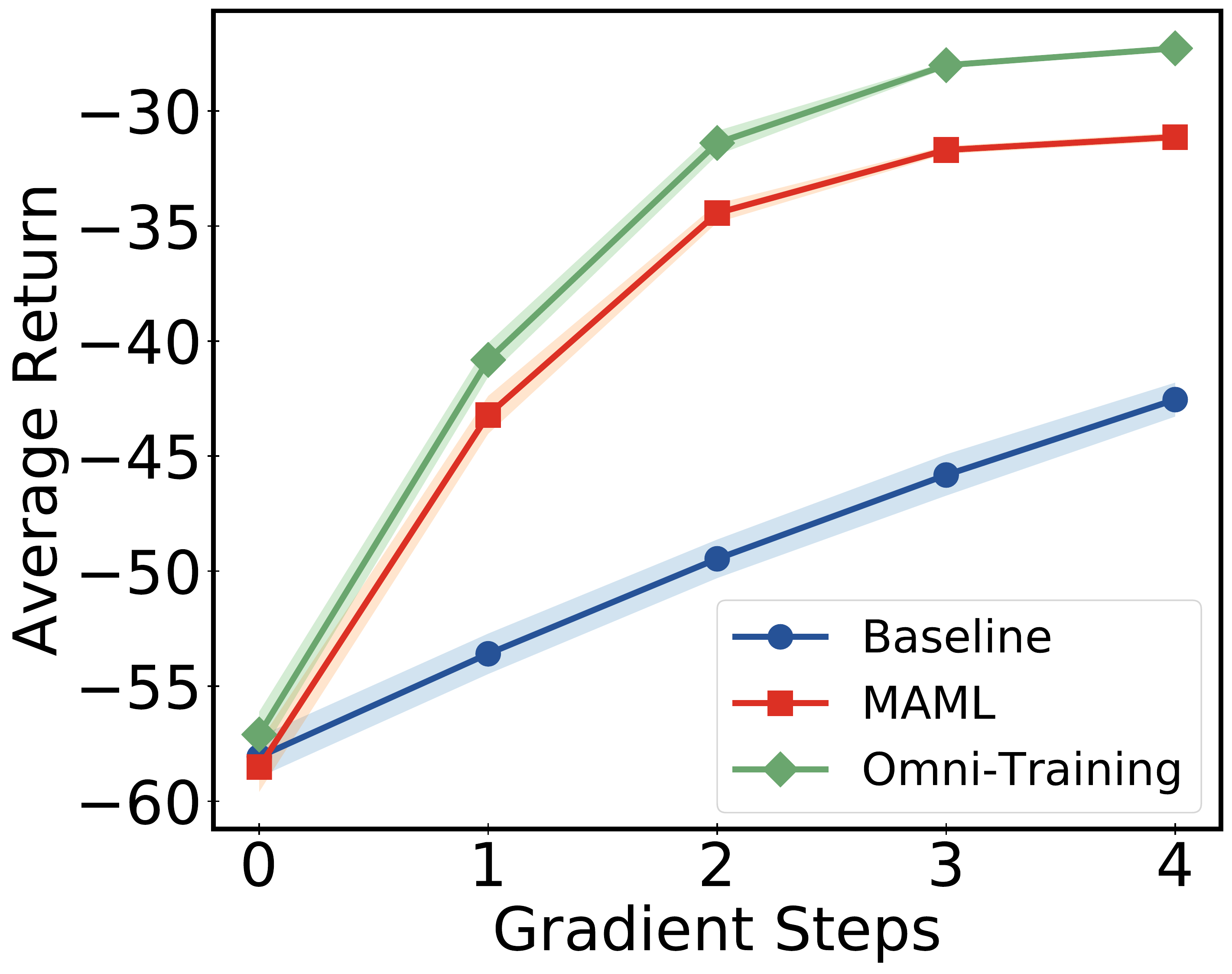} \label{fig:Reinforcement_Learning_2}}
	\hfil
	\subfloat[Locomotion-Direction]{
		\includegraphics[width=0.31\textwidth]{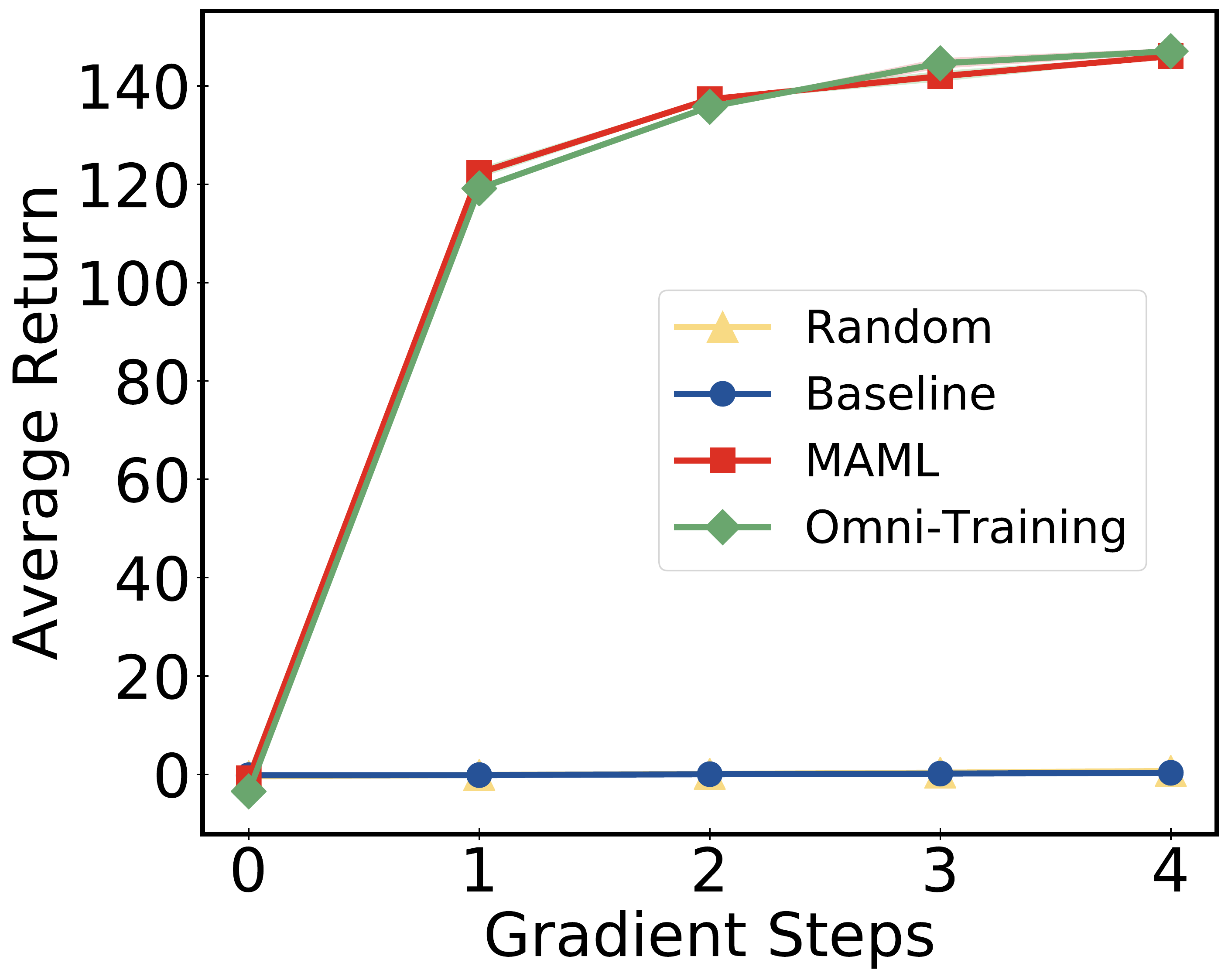} \label{fig:Reinforcement_Learning_3}}
	\hfil
	\caption{Average expected return of the tasks in the 2D Navigation environment and the Locomotion environment for reinforcement learning.}
	\label{fig:Reinforcement_Learning}
\end{figure*}

\subsection{Reinforcement Learning}

\myparagraph{Environments} For reinforcement learning problems, we follow the learning protocol in \cite{cite:ICML2017MAML} with several sets of tasks based on two simulated continuous control environments: \textbf{2D Navigation} and \textbf{Locomotion} in the \texttt{rllab} benchmark suite~\cite{cite:ICML16RLBenchmark}.

In the 2D Navigation environment, the goal is to move to a target position in 2D. The state space is the 2D location and the action space is the 2D velocity, where the action is in the range of $\left[-0.1, 0.1\right]$. The reward is the negative squared distance to the goal, and the episodes terminate when the agent is within $0.01$ of the goal or at the horizon of $H = 100$. We construct a task by randomly sampling a goal position from a unit square.

In the Locomotion environment, we adopt the agent in the \texttt{Mujoco} HalfCheetah environment~\cite{cite:IROS12Mujoco} and follow its state and action space. We evaluate on two sets of tasks. The first aims to run at a particular velocity. The reward is the negative absolute value between the current velocity and the goal velocity, which is chosen uniformly at random between $0.0$ and $2.0$ for different tasks. The second aims to run in a particular direction. The reward is the magnitude of the velocity in the forward or backward direction. The horizons of both tasks are set as $H = 200$.

\myparagraph{Implementation Details} We adopt the policy as a neural network with two fully-connected layers of $64$ hidden units and the Tanh activation function. We train the policy with the REINFORCE algorithm~\cite{cite:ML92Gradient}. We use the standard linear feature baseline proposed by \cite{cite:ICML16RLBenchmark}, which is fitted separately at each iteration for each sampled task in the batch. We train the model with $500$ iterations. In each iteration, $20$ different tasks are sampled for the 2D navigation environment and $40$ tasks are sampled for Locomotion, where $20$ trajectories are sampled for each task. During the test stage for few-shot reinforcement learning, we randomly sample $2000$ new tasks for evaluation. Each task contains trajectories with rewards as the support set. We use $20$ trajectories from each task for each gradient step and use $1$ to $4$ gradient steps for adaptation to new tasks. We use $20$ trajectories as the query set to compute the final testing reward of each task. We also use {Baseline}~\cite{cite:ICLR2019ACloserLook} as the pre-training method and \texttt{MAML}~\cite{cite:ICML2017MAML} as the meta-training method. In each iteration of meta-training, the policy is first trained using a single gradient step on the support trajectories with the inner loop step size $0.1$, and then meta-updated on the query trajectories with the outer loop step size $0.03$.

\myparagraph{Results} The results of the reinforcement learning tasks in the two environments are shown in Figure~\ref{fig:Reinforcement_Learning}. Omni-Training outperforms both Baseline and MAML with large margins in the 2D Navigation environment, which demonstrates that the model with both domain and task transferability can boost the generalization performance in this case. In the Locomotion environment, the performance gap between MAML and Baseline becomes larger, indicating more complex cross-task situations. Omni-Training still improves upon MAML in the velocity tasks. In the direction tasks, the pre-training method fails to generalize across these complex tasks with limited trajectories and updates, thereby performing similarly to the random initialization. In this extreme case, Omni-Training still performs comparably with MAML, without being negatively influenced. These results have proved the generalization ability of Omni-Training in a variety of complex situations.

\begin{figure*}[tbp]
	\centering
	\subfloat[mini-ImageNet$\rightarrow$CUB]{
		\includegraphics[width=0.31\textwidth]{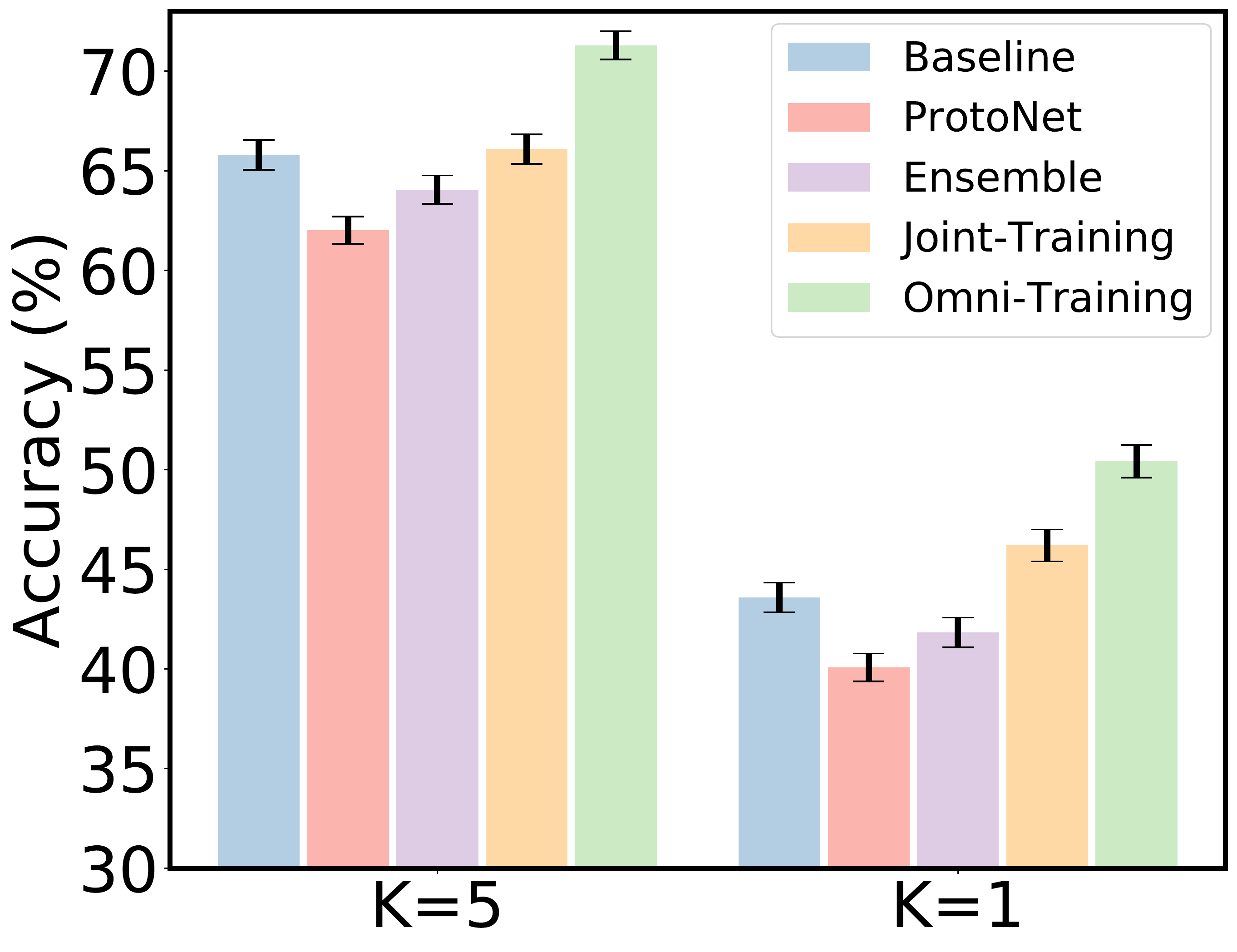}
		\label{fig:trivial1}
	}
	\subfloat[Regression]{
		\includegraphics[width=0.31\textwidth]{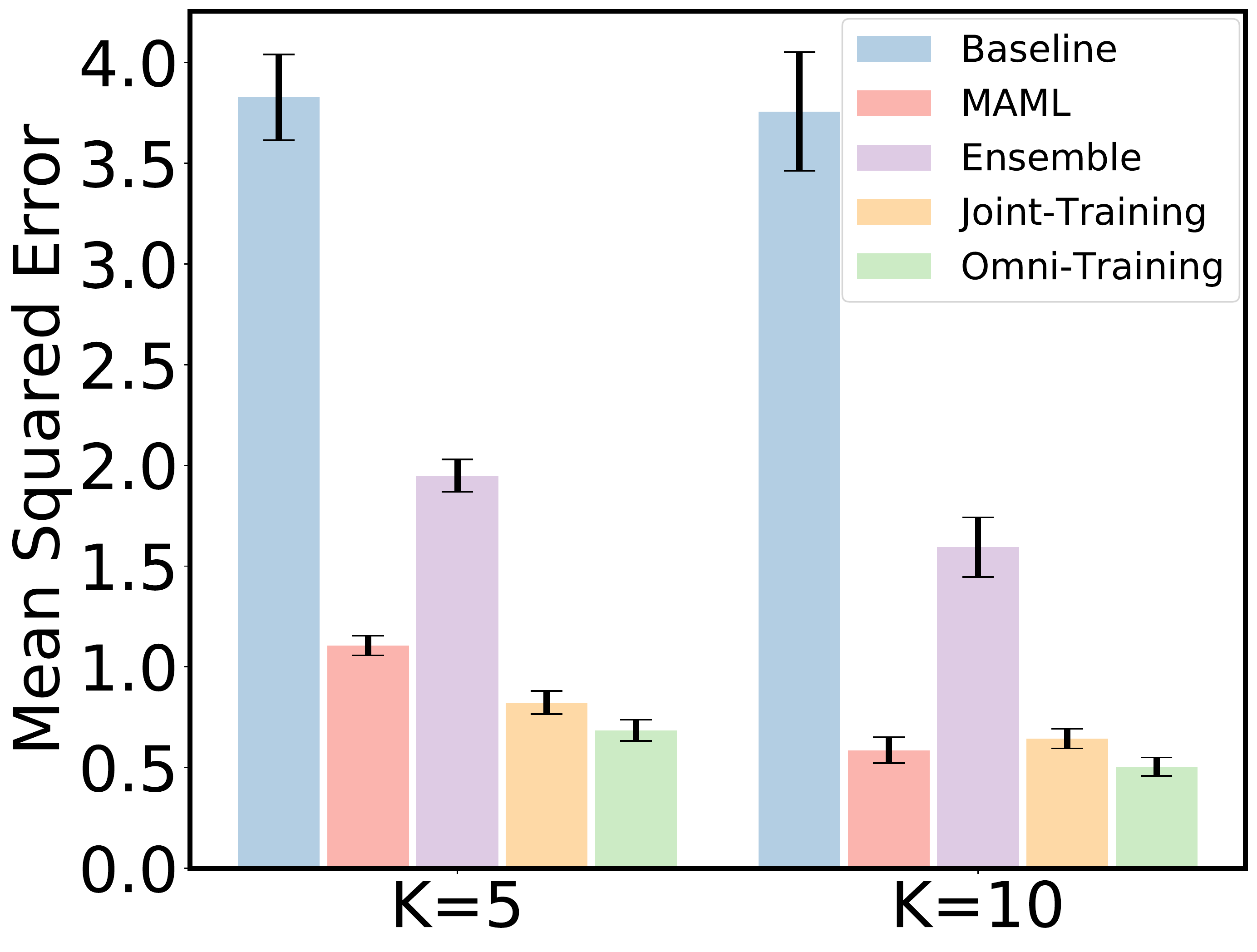}
		\label{fig:trivial2}
	}
	\subfloat[Extension]{
		\includegraphics[width=0.31\textwidth]{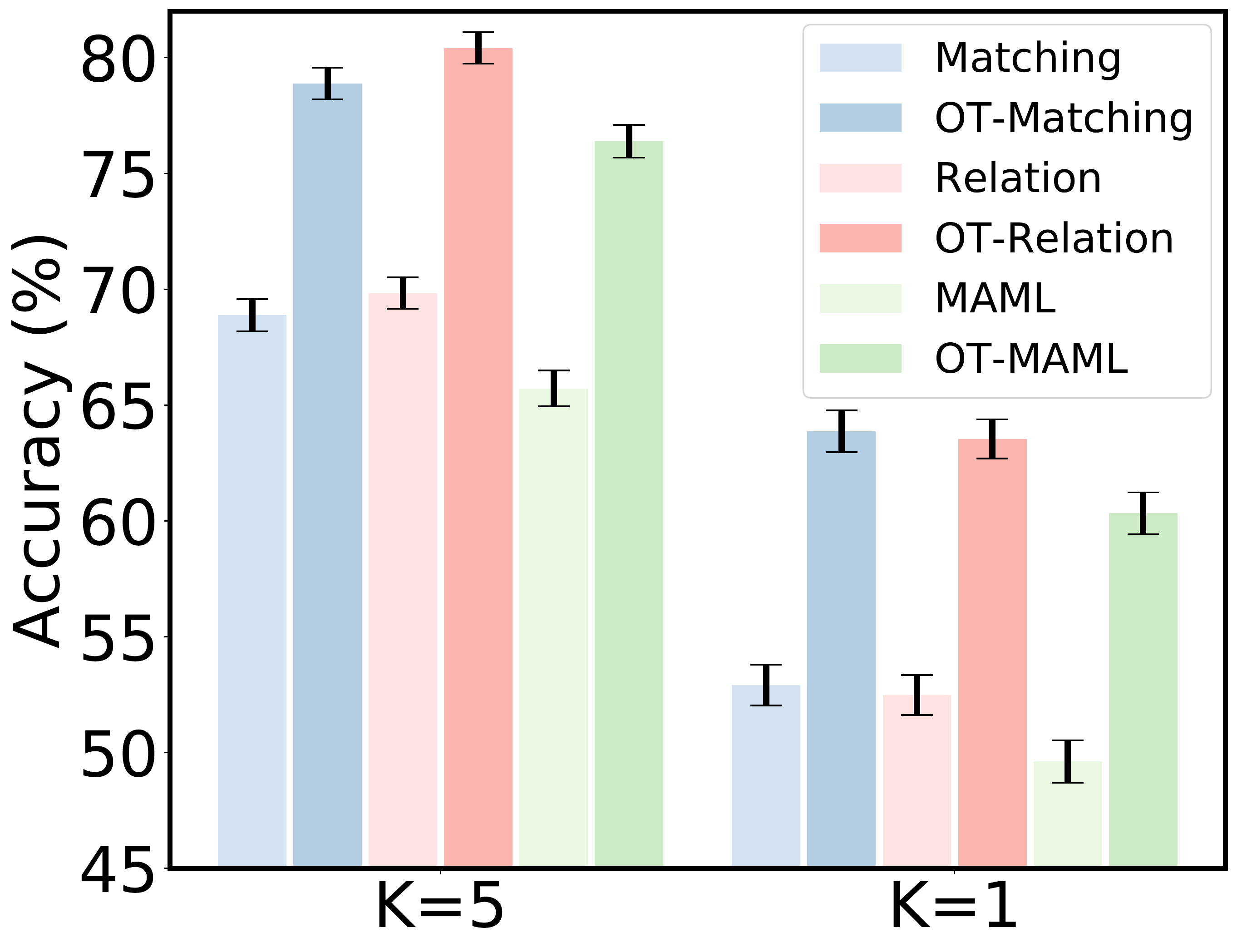}
		\label{fig:extension}
	}
	\caption{(a)(b) Comparison with two ways of simply combining pre-training and meta-training: Ensemble and Joint-Training. (c) Extension of the Omni-Training framework to other representation learning algorithms (OT: Omni-Training).}
\end{figure*}

\begin{figure*}[t]
	\subfloat[Pre-Training]{
		\includegraphics[width=0.32\textwidth]{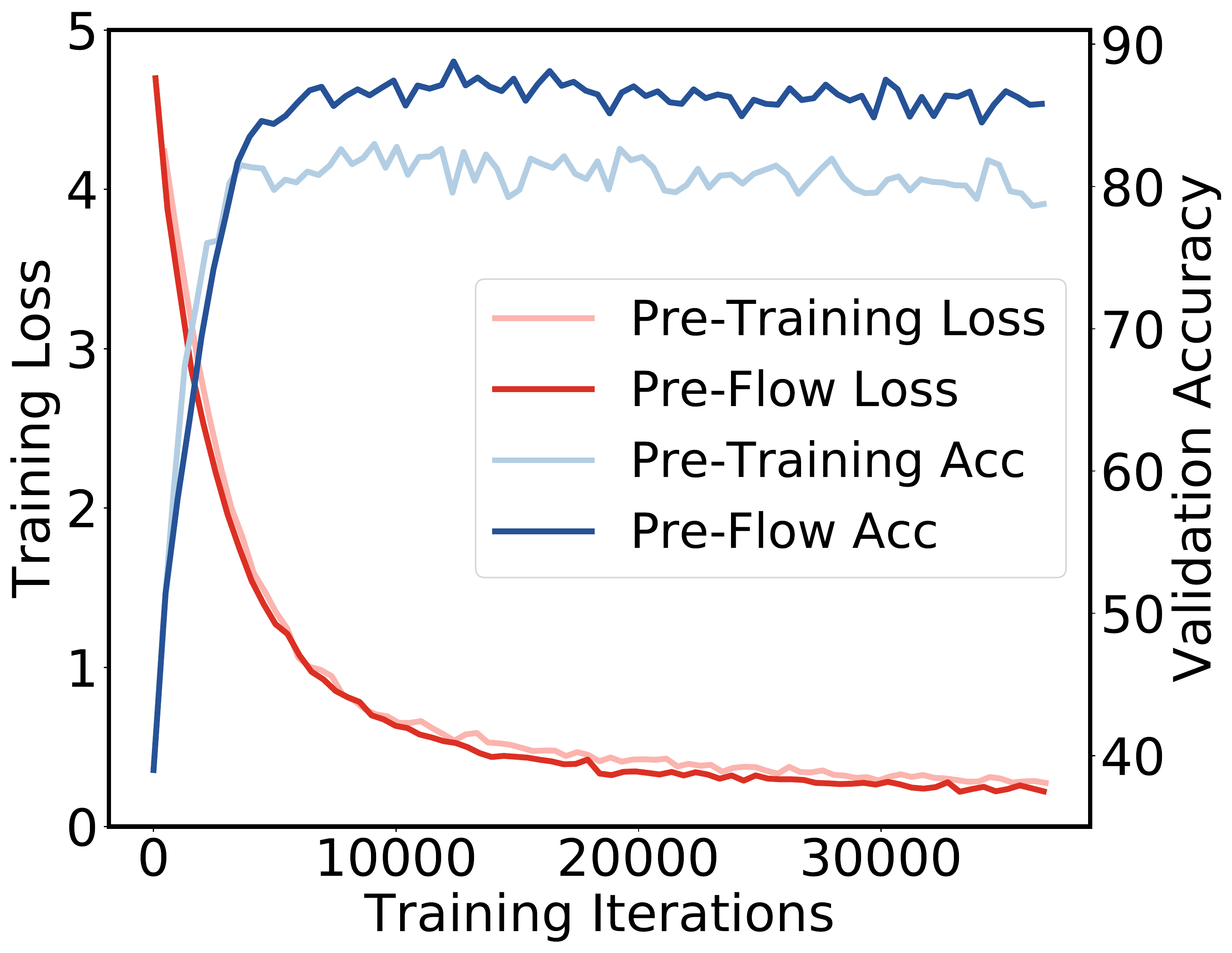}
		\label{fig:baselinetrain}
	}\hfil
	\subfloat[Meta-Training]{
		\includegraphics[width=0.33\textwidth]{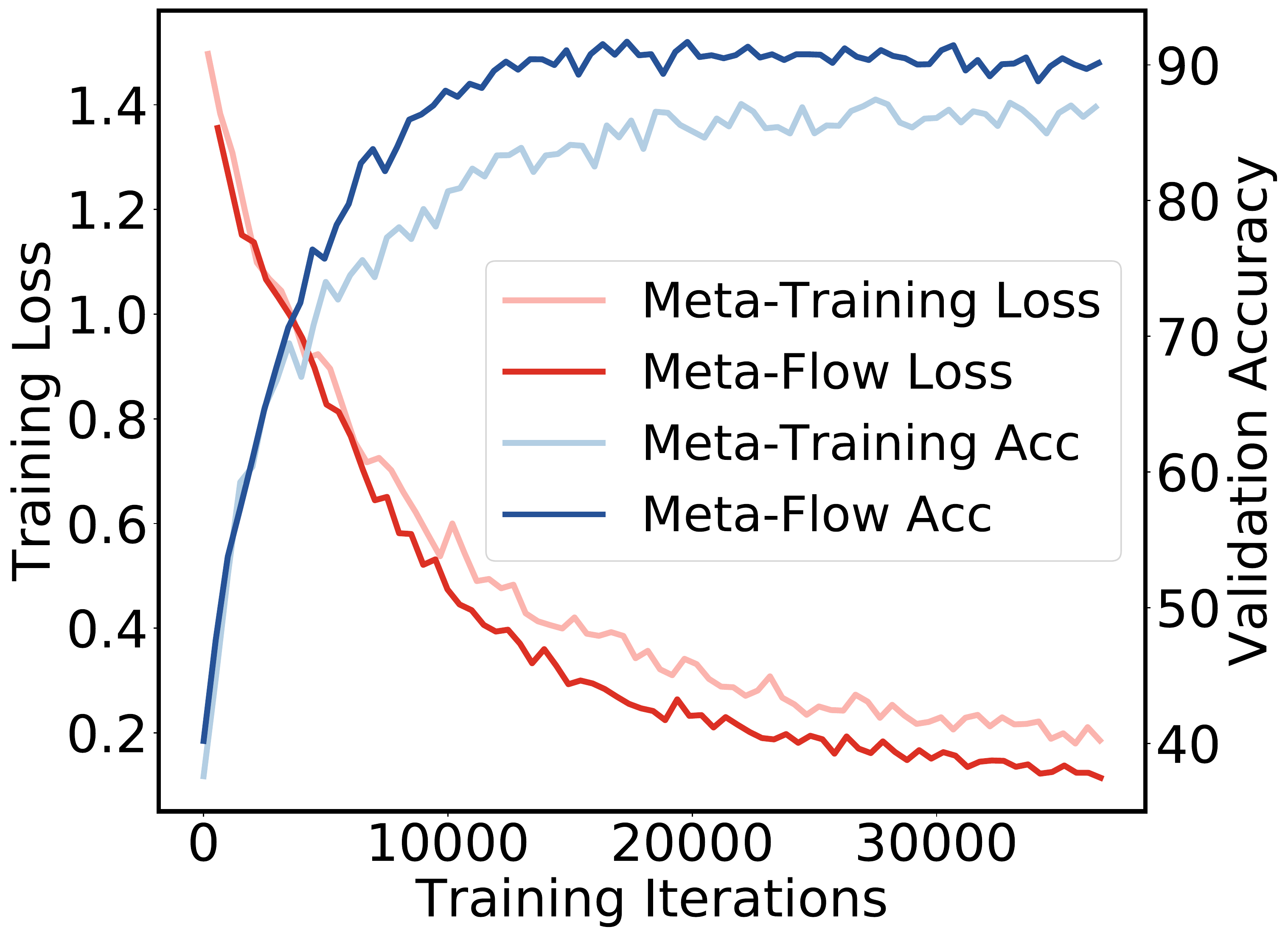}
		\label{fig:prototrain}
	}
	\subfloat[Transferability]{
		\includegraphics[width=0.30\textwidth]{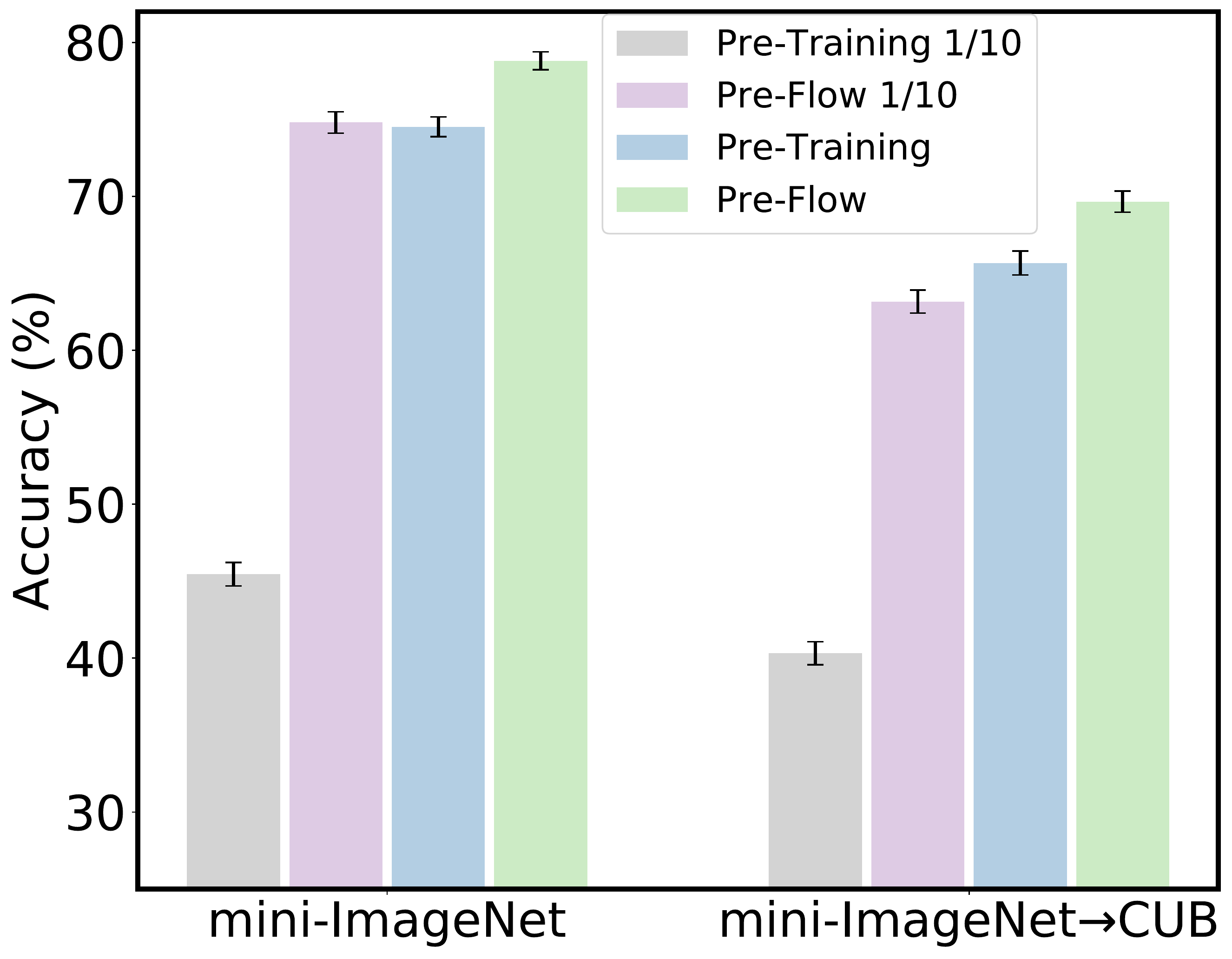}
		\label{fig:ablation_adaptability}
	}
	%\vskip -0.09in
	\caption{(a)(b) Training losses and validation accuracies of the pre-flow and meta-flow in Omni-Training, compared with their corresponding baselines; (c) The transferability of pre-training and pre-flow with different numbers of gradient steps.}
	%\vskip -0.16in
\end{figure*}

\section{Analysis}

In this section, we further empirically analyze and understand our proposed framework. Without specification, we use the ResNet-18 as the backbone. We use the Baseline in~\cite{cite:ICLR2019ACloserLook} as the pre-training method and the ProtoNet~\cite{cite:NIPS2017ProtoNet} as the meta-training method.

\subsection{Fine-grained Comparison with Baselines}
\myparagraph{Comparison with Simple Combinations} We compare Omni-Training with two simple combinations of pre-training and meta-training discussed in Section~\ref{sec:pre or meta}, \emph{i.e.} the ensemble of the two models trained separately (\texttt{Ensemble}) and joint-training with the losses of the two training paradigms (\texttt{Joint-Training}). We evaluate on the classification dataset mini-ImageNet$\rightarrow$CUB and the sinusoid regression dataset. We use $K=5$ or $K=1$ labeled samples in the support set in classification and use $K=5$ or $K=10$ labeled points with $2$ gradient steps of parameter update in regression. As shown in Figure~\ref{fig:trivial1} and~\ref{fig:trivial2}, Ensemble and Joint-Training do not always lead to improvements, and the performance gain is minor. Omni-Training instead outperforms all the compared methods consistently, which demonstrates that the proposed Omni-Net and the Omni-Loss designs provide a better solution to bridge pre-training and meta-training and acquire both domain transferability and task transferability.

\myparagraph{Extension to Other Algorithms} Despite the competitive performance on various benchmarks, we also want to demonstrate that different few-shot learning algorithms can benefit from the Omni-Training framework. We extend Omni-Training to more algorithms. Since most pre-training algorithms adopt the similar pre-training and fine-tuning process, we mainly investigate the varieties of meta-training algorithms including MatchingNet~\cite{cite:NIPS2016MatchingNet}, MAML~\cite{cite:ICML2017MAML} and RelationNet~\cite{cite:CVPR2018RelationNet}. We conduct experiments in the mini-ImageNet dataset since some algorithms cannot deal with the regression problem. As shown in Figure~\ref{fig:extension}, Omni-Training with different algorithms significantly outperforms the corresponding baselines. This demonstrates that our framework can generally accommodate different few-shot learning algorithms.

\begin{figure*}[ht]
	\centering
	\subfloat{
		\includegraphics[width=0.90\textwidth]{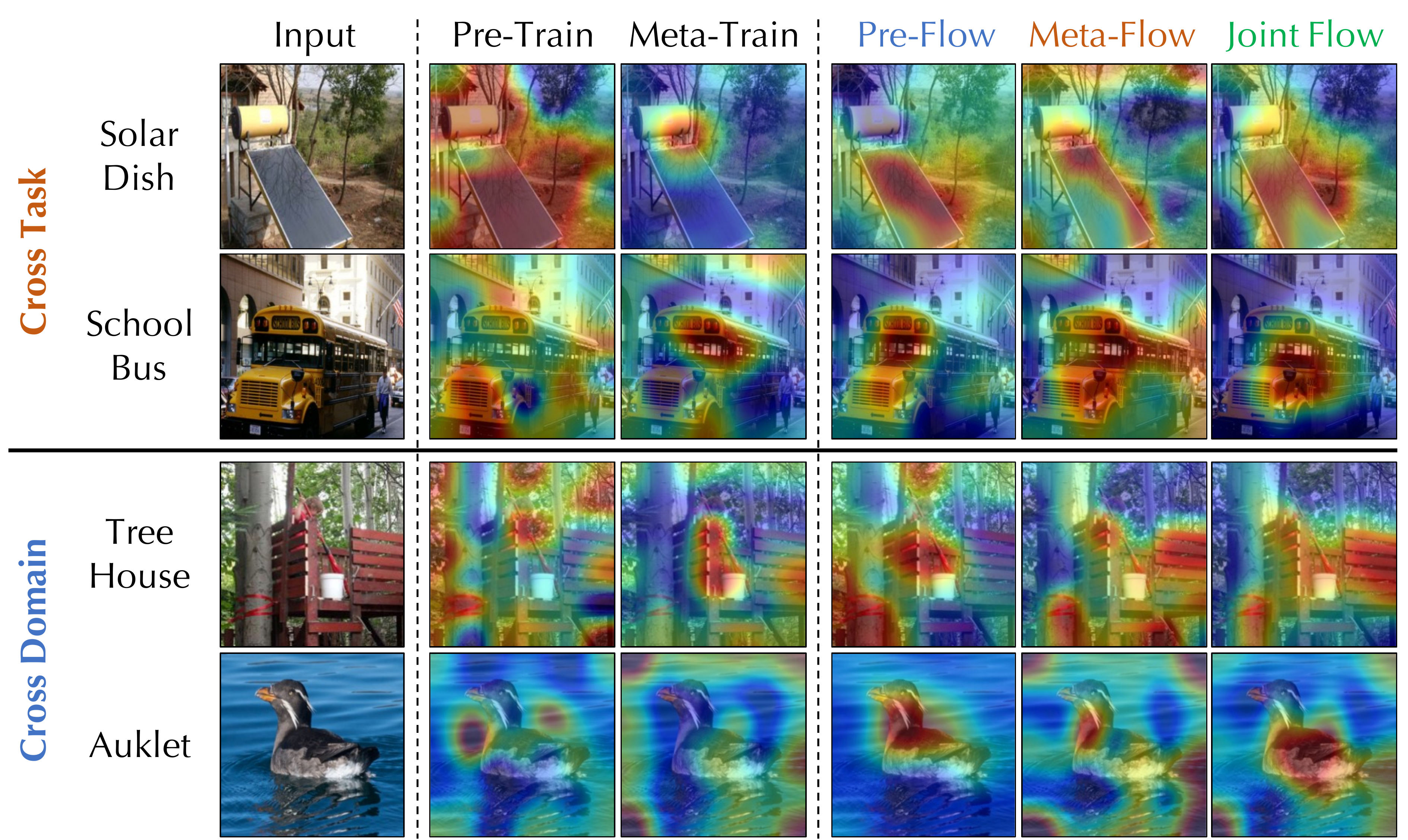}
		\label{fig:omni_attention}}
	\hfil
	\caption{Attention comparison of the representations from the pre-training model, the meta-training model and the three data flows in Omni-Training.}\label{fig:omni_attention_study}
\end{figure*}

\myparagraph{Comparison of Each Flow with Baselines} We investigate whether the coordination of pre-training and meta-training with the shared parameters in our \emph{tri-flow} architecture can improve the performance of specific flows. Figure~\ref{fig:baselinetrain} reports the training losses and validation accuracies of the \emph{pre-flow} in Omni-Training and pre-training algorithm {Baseline}~\cite{cite:ICLR2019ACloserLook} alone, while Figure~\ref{fig:prototrain} reports the results of the \emph{meta-flow} in Omni-Training and the meta-training algorithm {ProtoNet}~\cite{cite:NIPS2017ProtoNet}. The experiments are conducted in the CUB dataset with $K=5$. The pre-flow and the meta-flow in Omni-Training reach lower losses and higher accuracies than the baselines trained independently. Even though the pre-flow and Baseline achieve nearly the same training loss, the pre-flow achieves much higher validation accuracy than Baseline. This shows that the knowledge communication enables pre-flow to obtain part of task transferability and meta-flow to obtain part of domain transferability to improve their performance.

We also compare the transferability of the pre-training method and the pre-flow on the mini-ImageNet and mini-ImageNet$\rightarrow$CUB datasets. As shown in Figure~\ref{fig:ablation_adaptability}, the pre-flow also outperforms pre-training in various situations. We further investigate fine-tuning the representations with $1/10$ gradient steps. The performance of the pre-training model drops a lot with limited updates, but the pre-flow still performs well and comparably with the pre-training model updated $10$ more times. This reveals that the pre-flow also acquires task transferability to fast adapt across tasks. These results demonstrate that Omni-Training coordinates the two parallel flows and makes each gain the other kind of transferability.

\begin{figure*}[tbp]
	\centering
	\subfloat[CUB]{
		\includegraphics[width=0.315\textwidth]{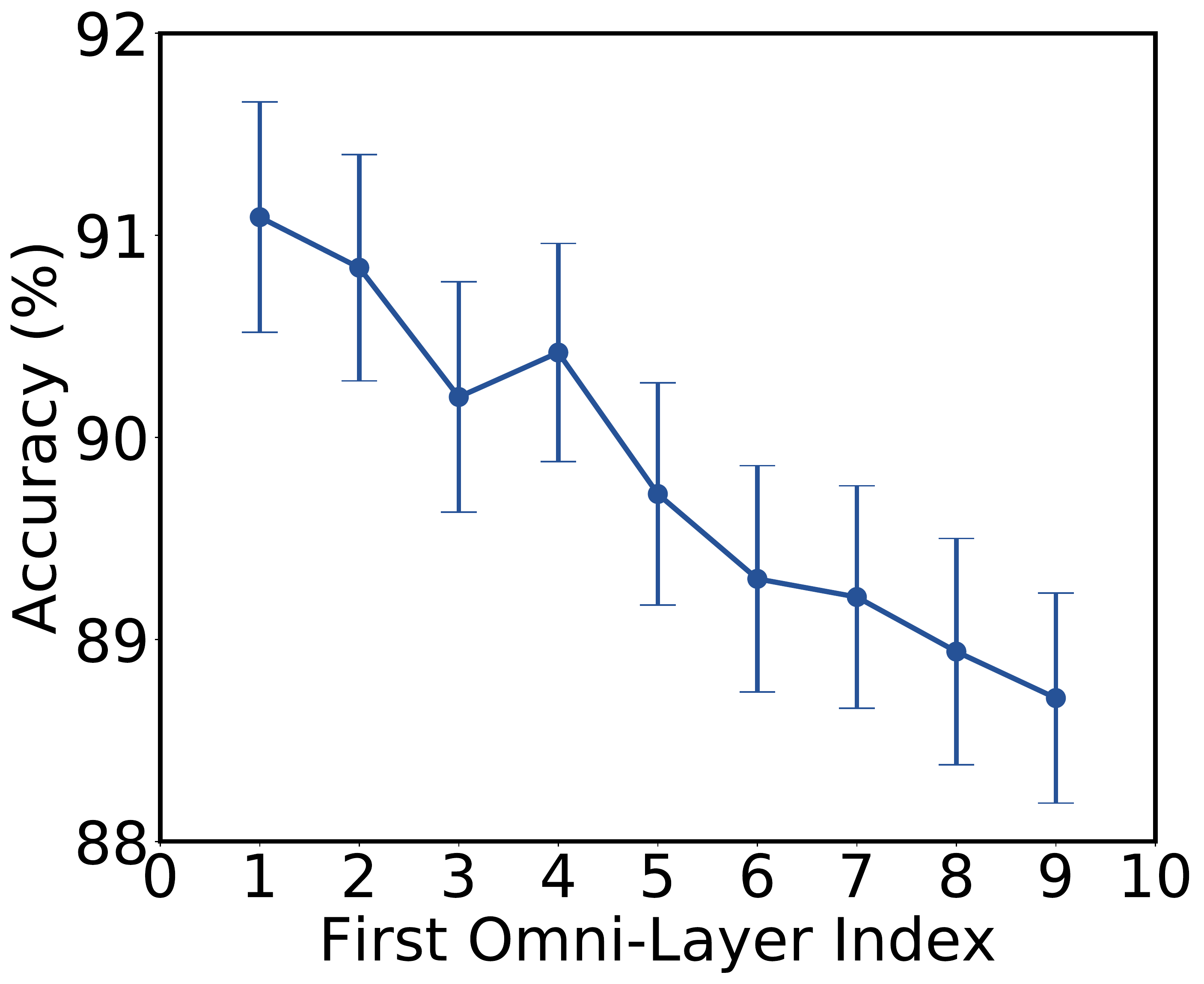}
		\label{fig:startlayer1}
	}\hfil
	\subfloat[mini-ImageNet$\rightarrow$CUB]{
		\includegraphics[width=0.315\textwidth]{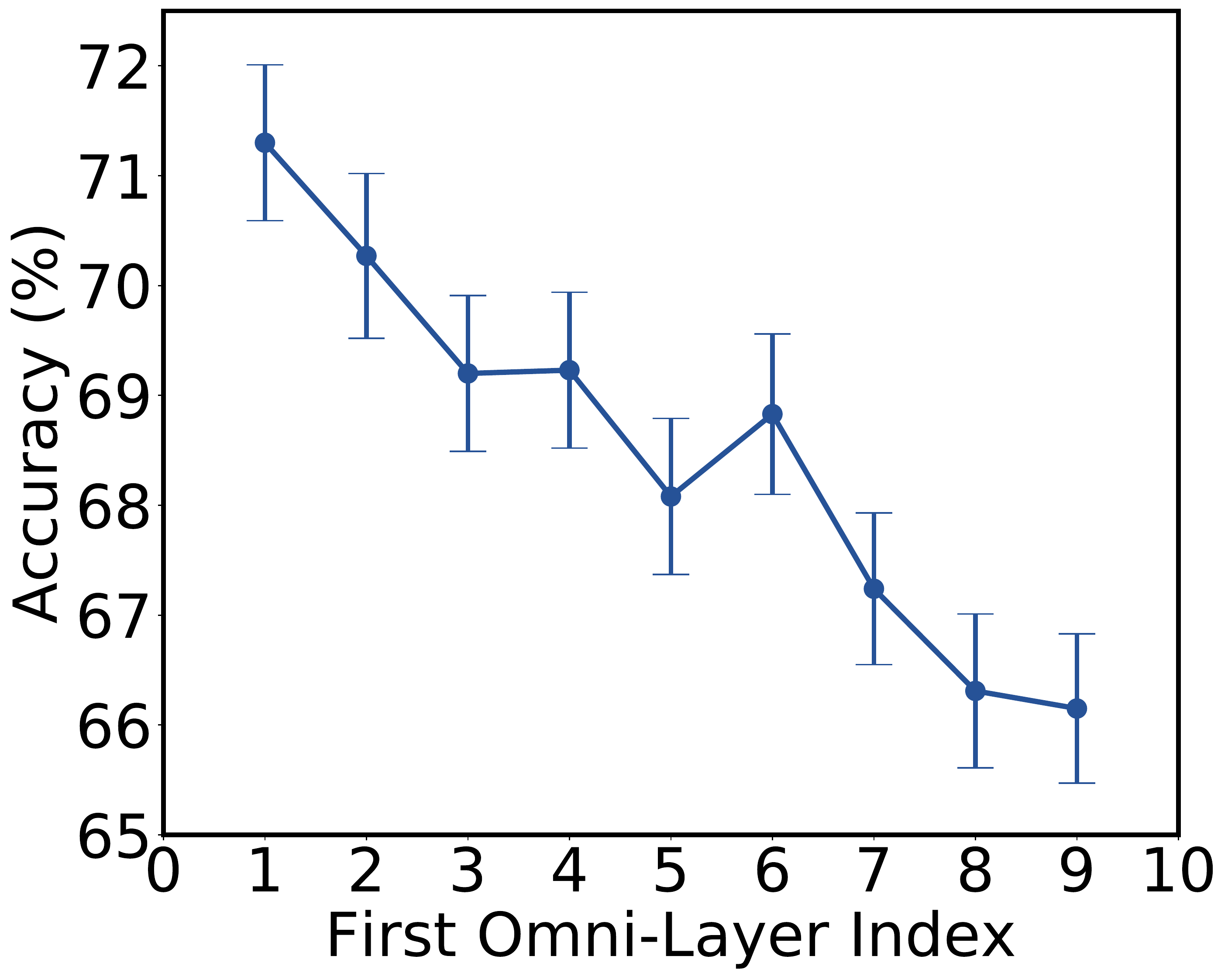}
		\label{fig:startlayer2}
	}
	\subfloat[Parameter Sensitivity]{
		\includegraphics[width=0.315\textwidth]{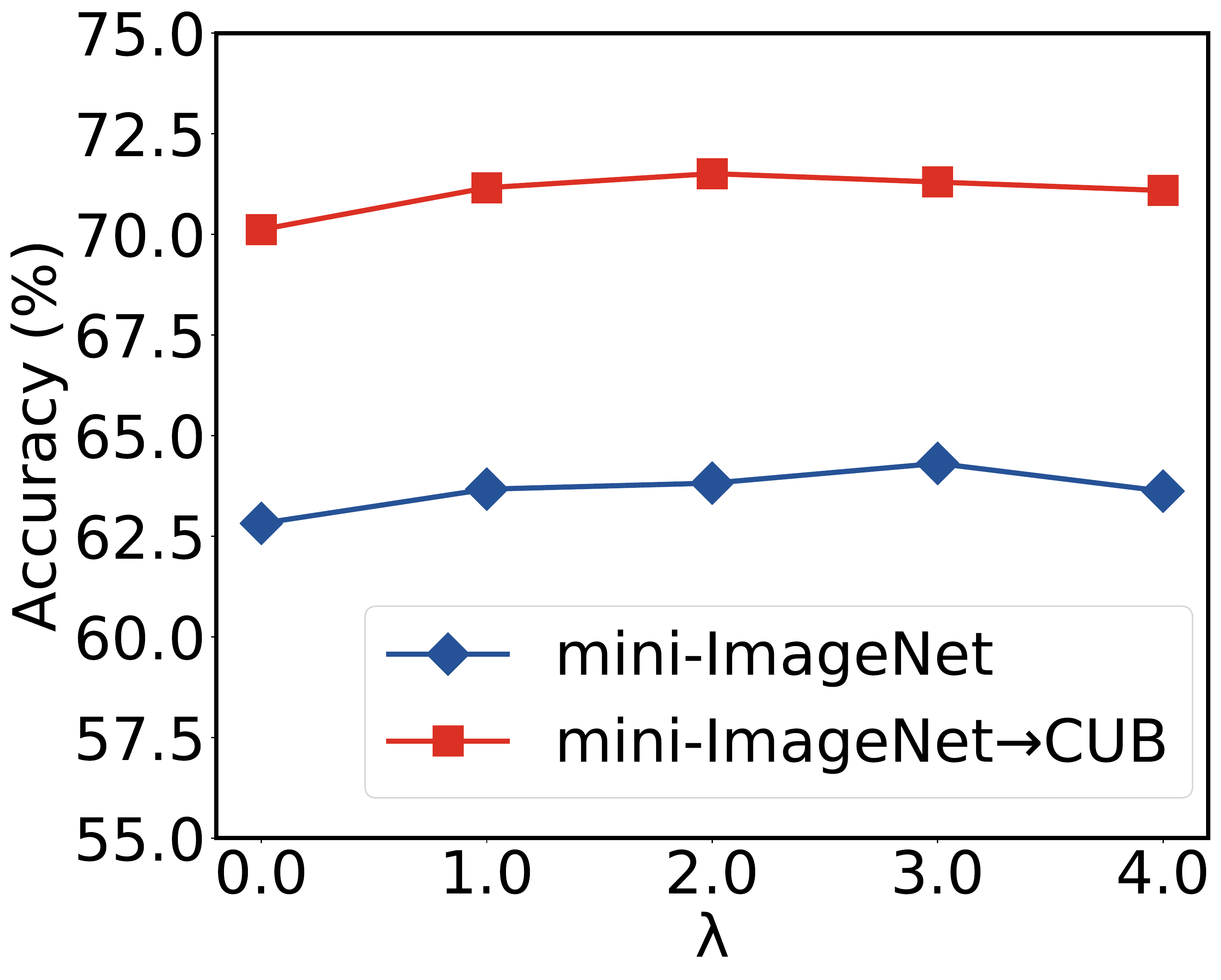}
		\label{fig:sensitivity}
	}
	\caption{(a)(b) Results of modifying different numbers of layers into Omni-Layers on the CUB dataset and the mini-ImageNet$\rightarrow$CUB dataset respectively. (c) The sensitivity of the performance with respect to the hyper-parameters $\lambda$.}
\end{figure*}

\begin{table}[t]
    \addtolength{\tabcolsep}{-2pt}
	\begin{center}
		\caption{Ablation study on the losses in the Omni-Training framework.}
		\label{table:ablation}
		%\vskip 0.05in
% 		\resizebox{0.7\columnwidth}{!}{%
			%\begin{Huge}
			%\begin{sc}
			%\renewcommand\tabcolsep{1.5pt}
			\begin{tabular}{ccccccc}
				\toprule
				% 	\hline
				$\mathcal{L}_\texttt{pre}$ & $\mathcal{L}_\texttt{meta}$ & $\mathcal{L}_\texttt{joint}$ & $\mathcal{R}$ & {ImageNet} & {CUB}   & {ImageNet$\rightarrow$CUB} \\
				\midrule
				% 	\hline
				\ymark                     & -                           & -                            & -                                                    & $74.37$         & $82.80$ & $65.57$                         \\
				-                          & \ymark                      & -                            & -                                                     & $73.66$         & $87.91$ & $62.14$                         \\
				-                          & -                           & \ymark                       & -                                                    & $76.95$         & $85.15$ & $66.09$                         \\
				\ymark                     & \ymark                      & -                            & -                                                     & $78.12$         & $87.34$ & $69.66$                         \\
				\ymark                     & \ymark                      & \ymark                       & -                                                    & $80.79$         & $88.54$ & $70.12$                         \\
				\ymark                     & \ymark                      & \ymark                       & \ymark                                           & $81.26$         & $91.09$ & $71.30$                         \\
				\bottomrule
				% 	\hline
			\end{tabular}%
			%\end{sc}
			%\end{Huge}
% 		}
	\end{center}
	%\vskip -0.21in
\end{table}

\begin{table}[t]
	\begin{center}
		\caption{Ablation study on the model size.}
		\label{table:Parameters}
		%		\resizebox{0.48\textwidth}{!}{%
		\begin{tabular}{lcccc}
			\toprule
			% \hline
			Method        & \#Params & ImageNet & CUB     & ImageNet$\rightarrow$CUB \\
			\midrule
			% \hline
			ProtoNet      & 11.17M   & $73.68$  & $87.42$ & $62.02$                  \\
			ProtoNet*     & 13.98M   & $73.44$  & $87.81$ & $61.27$                  \\
			Omni-Training & 13.98M   & $81.26$  & $91.09$ & $71.30$                  \\
			\bottomrule
			% \hline
		\end{tabular}%
		%		}
		%}
		%\end{sc}
		%\end{small}
	\end{center}
\end{table}

\myparagraph{Comparison of Attention Maps} We compare the spatial attention in different representations learned by pre-training, meta-training and the three data flows in Omni-Training. From Figure~\ref{fig:omni_attention_study}, we observe that pre-training representations focus on a broad area containing the objects as well as some noisy context, which fully grab the domain knowledge but lack some concentration on the important information to discriminate different categories. On the contrary, the meta-training representations focus on a very small area with very concise information, which is easy to generalize across tasks quickly but also easy to make mistakes when the attending area deviates only a little from the objects. Such deviation is more likely to occur with the domain shift. Such attention heatmaps are consistent with our analyses before that pre-training learns representations with higher domain transferability while meta-training learns representations with higher task transferability.

Switching to Omni-Training, the pre-flow focuses on a more concise area only including the whole object while ignoring the noisy context. The meta-flow focuses on a broader area to grab more knowledge in the whole domain and increase the tolerance of mistakes. This observation demonstrates that there is knowledge transfer between pre-flow and meta-flow, which coordinates these two flows and improves them with the other kind of transferability. The joint-flow shows a different attention map from the pre-flow and the meta-flow. This also demonstrates that the three flows in the Omni-Training framework focus on different areas on the input space and form a more comprehensive understanding of the datapoints.

\subsection{Framework Analysis}

\myparagraph{Ablation Study of Losses} We conduct an ablation study by using different combinations of losses in the Omni-Training framework. For the losses of $\mathcal{L}_\texttt{pre}$, $\mathcal{L}_\texttt{meta}$ and $\mathcal{L}_\texttt{joint}$, if we do not use any of the three losses, we will not use the corresponding branch for inference. We report results on mini-ImageNet, CUB and mini-ImageNet$\rightarrow$CUB datasets with $K=5$ in Table~\ref{table:ablation}. We observe that all of the loss functions in the tri-flow design including the self-distillation regularization contribute to the improvement of the Omni-Training framework.

\myparagraph{Influence of the Model Size} In Omni-Net, we use lightweight $1\times1$ convolution layers for the parallel branches. Although the number of parameters does not increase significantly (from $11.17$M to $13.98$M if we use ResNet-18), there is still a concern that the performance gain of Omni-Training may come from the increase in the model size. Thus, we add the same parameters as these additional $1\times1$ convolution layers to the original ResNet-18 backbone, and denote it as ResNet-18*. Though having the same number of parameters, ResNet-18* is different from our Omni-Training backbone because it does not have different data flows inside respectively for pre-training and meta-training, and is only trained with one learning paradigm. We train ProtoNet~\cite{cite:NIPS2017ProtoNet} with the ResNet-18* backbone (denoted as ProtoNet*) and report the accuracy with the support set size $K=5$ in Table~\ref{table:Parameters}.

Despite having more parameters, ProtoNet* does not show obvious improvement over ProtoNet. This indicates that simply increasing the model complexity does not ensure better performance. Omni-Training has comparable parameters with ProtoNet*, but outperforms ProtoNet* with a large margin. This reveals that the main reason that improves the performance is not increasing the model size, but coordinating pre-training and meta-training to learn deep representations with both domain transferability and task transferability.

\myparagraph{Backbone Modification} We investigate the incluence of the number of Omni-Layers of the backbone. Since ResNet-18 is composed of $8$ Res-Blocks, we attempt to keep the first $n$ Res-Blocks unchanged and transform the rest $8-n$ blocks into Omni-Layers. The first index of the block with Omni-Layers is $n+1$. We train the models with these modified backbones. We report classification results with $K=5$ in the CUB dataset (Figure~\ref{fig:startlayer1}) and the mini-ImageNet$\rightarrow$CUB dataset (Figure~\ref{fig:startlayer2}). When the index of the first block with Omni-Layers is $1$, which means the whole backbone is changed into Omni-Net, the model performs best. As the index increases, which means more preceding layers are completely shared between different flows as done in Multi-Task Learning, the accuracy drops sharply. This reveals the efficacy of the Omni-Layers on learning the three flows to coordinate pre-training and meta-training. Omni-Net is a general-purpose backbone for few-shot learning.

\myparagraph{Parameter Sensitivity} We analyze the sensitivity of the loss trade-off hyper-parameter $\lambda$. We report the accuracy on the mini-ImageNet dataset with $K=1$ and on the cross-domain mini-ImageNet$\rightarrow$CUB dataset with $K=5$ in Figure~\ref{fig:sensitivity}. We observe that the model performs well in a range of parameters: $[1.0,3.0]$. However, the performance degrades when setting $\lambda=0$, \emph{i.e.}, removing the self-distillation regularization. In general, we use the same hyper-parameter: $\lambda=3.0$ for the different tasks in our experiments to avoid over-tuning it.

\section{Conclusion}
This paper focuses on learning transferable representations for few-shot learning, which enables the model to fast generalize to new domains and tasks with a few examples. We pinpoint that domain transferability and task transferability are the key factors to data-efficiency in downstream tasks. We further empirically show that pre-training and meta-training methods and simple combinations of them cannot obtain both domain transferability and task transferability, so we propose Omni-Training to bridge pre-training and meta-training with both types of transferability.
With the tri-flow Omni-Net architecture, the model preserves the specific transferability of pre-training and meta-training and coordinates these flows by routing their representations via the joint-flow, making each gain the other kind of transferability. We design an Omni-Loss to learn the three flows and impose a self-distillation regularization to enable knowledge transfer across the training process. Omni-Training is a general framework that accommodates various existing pre-training and meta-training algorithms. Thorough evaluations on cross-task and cross-domain datasets in classification, regression and reinforcement learning problems shows that Omni-Training consistently and clearly outperforms the state-of-the-art deep learning methods for few-shot learning.

\ifCLASSOPTIONcompsoc
  \section*{Acknowledgments}
\else
  \section*{Acknowledgment}
\fi

This work was supported by the National Megaproject for New Generation AI  (2020AAA0109201), National Natural Science Foundation of China (62022050 and 62021002), Beijing Nova Program (Z201100006820041), and BNRist Innovation Fund (BNR2021RC01002).

\ifCLASSOPTIONcaptionsoff
  \newpage
\fi

\bibliographystyle{IEEEtran}
\bibliography{omni}

\end{document}